%% file: main.tex
\documentclass[sigconf]{acmart}

\usepackage{booktabs}
\usepackage{subfigure}
\usepackage{caption}
\usepackage{enumitem}
\usepackage{balance}

\usepackage{bm}
\usepackage{rotating}
\usepackage{amsmath,amssymb,amsfonts}

\usepackage{multirow}
\usepackage{booktabs}
\usepackage{algorithm}
\usepackage{algorithmic}

\newtheorem{lemma}{Lemma}

\AtBeginDocument{%
  }

\setcopyright{acmcopyright}
\copyrightyear{2023}
\acmYear{2023}
\setcopyright{acmlicensed}\acmConference[MM '23]{Proceedings of the 31st
ACM International Conference on Multimedia}{October 29-November 3,
2023}{Ottawa, ON, Canada}
\acmBooktitle{Proceedings of the 31st ACM International Conference on
Multimedia (MM '23), October 29-November 3, 2023, Ottawa, ON, Canada}
\acmPrice{15.00}
\acmDOI{10.1145/3581783.3612264}
\acmISBN{979-8-4007-0108-5/23/10}

\acmSubmissionID{2461}

\begin{document}

\def\method{SA-GDA}
\title{\method{}: Spectral Augmentation for Graph Domain Adaptation}


\author{Jinhui Pang}
\affiliation{%
  \institution{Beijing Institute of Technology}
  \city{Beijing}
  \country{China}}
\email{pangjinhui@bit.edu.cn}

\author{Zixuan Wang}
\affiliation{%
  \institution{Beijing Institute of Technology}
  \city{Beijing}
  \country{China}}
\email{3220211081@bit.edu.cn}

\author{Jiliang Tang}
\affiliation{%
  \institution{Michigan State University}
  \city{East Lansing}
  \state{Michigan}
  \country{USA}}
\email{tangjili@msu.edu}

\author{Mingyan Xiao}
\affiliation{%
  \institution{California State Polytechnic University, Pomona}
  \city{Pomona}
  \state{California}
  \country{USA}}
\email{mxiao@cpp.edu}

\author{Nan Yin}
\authornote{Corresponding author.}
\affiliation{%
  \institution{Mohamed bin Zayed University of Artificial Intelligence}
  \city{Abu Dhabi}
  \country{UAE}}
\email{nan.yin@mbzuai.ac.ae}



\renewcommand{\shortauthors}{Jinhui Pang, Zixuan Wang, Jiliang Tang, Mingyan Xiao, \& Nan Yin} %% No italics

% \begin{abstract}
%   A clear and well-documented \LaTeX\ document is presented as an
%   article formatted for publication by ACM in a conference proceedings
%   or journal publication. Based on the ``acmart'' document class, this
%   article presents and explains many of the common variations, as well
%   as many of the formatting elements an author may use in the
%   preparation of the documentation of their work.
% \end{abstract}






\input{1.abstract}

\maketitle

\input{2.introduction}
\input{3.related}

\input{4.method}

\input{5.experiment}

\input{6.conclusion}
\input{8.Acknowledge}

\newpage

\bibliographystyle{ACM-Reference-Format}
\balance
\bibliography{sample-base}

\appendix
\input{7.Appendix}

\end{document}

%% file: 1.abstract.tex
\begin{abstract}

Graph neural networks (GNNs) have achieved impressive impressions for graph-related tasks. However, most GNNs are primarily studied under the cases of signal domain with supervised training, which requires abundant task-specific labels and is difficult to transfer to other domains. 
There are few works focused on domain adaptation for graph node classification. They mainly focused on aligning the feature space of the source and target domains, without considering the feature alignment between different categories, which may lead to confusion of classification in the target domain. However, due to the scarcity of labels of the target domain, we cannot directly perform effective alignment of categories from different domains, which makes the problem more challenging.
In this paper, we present the \textit{Spectral Augmentation for Graph Domain Adaptation (\method{})} for graph node classification. First, we observe that nodes with the same category in different domains exhibit similar characteristics in the spectral domain, while different classes are quite different. Following the observation, we align the category feature space of different domains in the spectral domain instead of aligning the whole features space, and we theoretical proof the stability of proposed \method{}. 
Then, we develop a dual graph convolutional network to jointly exploits local and global consistency for feature aggregation. Last, we utilize a domain classifier with an adversarial learning submodule to facilitate knowledge transfer between different domain graphs. 
Experimental results on a variety of publicly available datasets reveal the effectiveness of our \method{}. 
\end{abstract}

%% file: 2.introduction.tex
\section{Introduction}

Graphs are more and more popular due to their capacity to represent structured and relational data in a wide range of fields~\cite{dai2021hyperbolic,xu2021discrimination}. 
As a basic problem, graph node classification aims to predict the category of each node and has already been widely discussed in various fields, such as protein-protein interaction networks~\cite{gao2023hierarchical,fout2017protein}, citation networks~\cite{kipf2017semi,hamilton2017inductive}, and time series prediction. They typically follow the message-passing mechanisms and learn the discriminative node representation for classification.

\begin{figure}[t]
\centering
\includegraphics[width=8cm,keepaspectratio=true]{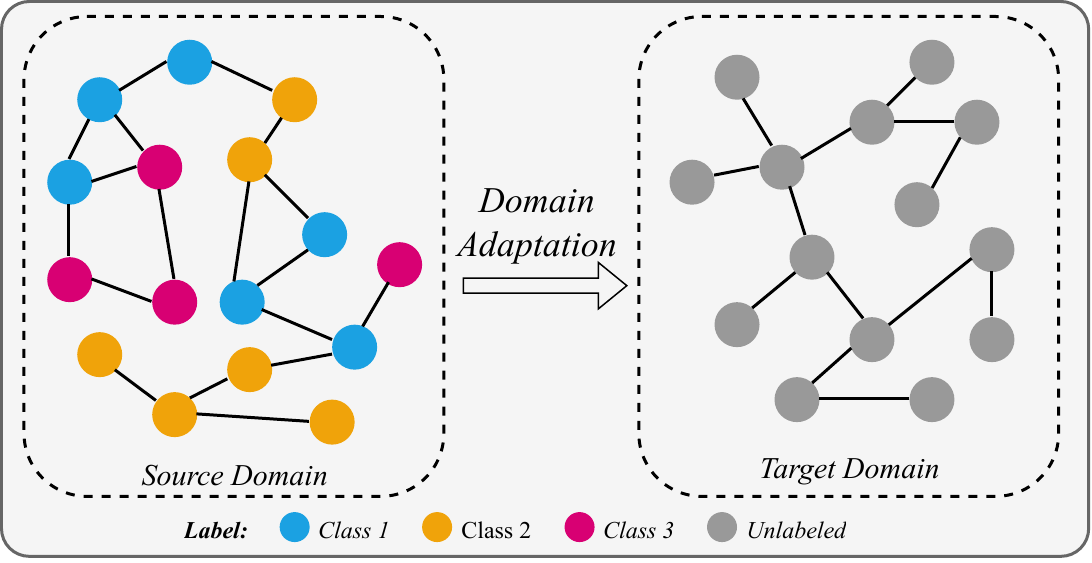}
\caption{Illustration of graph domain adaptation. The source graph is all labeled and the target graph is unlabeled, domain adaptation aims to train a classifier in the source domain so as to transfer to the target domain classification.}
\label{fig:da}
\vspace{-0.2cm}
\end{figure}

Although existing GNNs exhibit impressive performance, they typically rely on supervised training methods~\cite{xu2019powerful,ju2024survey}, which necessitates a large amount of labeled nodes. Unfortunately, labeling can be a costly and time-consuming process in many scientific domains~\cite{hao2020asgn,suresh2021adversarial}. Additionally, annotating graphs in certain disciplines often requires specialized domain knowledge, which prevents a well-trained model from being transferable to a new graph. To address this issue, we investigate the problem of unsupervised domain adaptation for graph node classification, which utilizes labeled source data and unlabeled target data to accomplish accurate classification on the target domain. The illustration of unsupervised graph domain adaptation is shown in Figure~\ref{fig:da}.

There are limited works attempting to apply domain adaptation for graph-structure data learning~\cite{shen2020adversarial,dai2019network,2020Unsupervised}. CDNE~\cite{shen2020adversarial} learns the cross-domain node embedding by minimizing the maximum mean discrepancy (MMD) loss, which cannot jointly model graph structures and node attributes, limiting the modeling capacity. AdaGCN~\cite{dai2019network} uses GNN as a feature extractor to learn node representations and utilizes adversarial learning to learn domain invariant node representation. 
UDA-GCN~\cite{2020Unsupervised} utilizes the random walk method to capture the local and global information to enhance the representation ability of nodes.
Though these methods have achieved good results, there are still two fundamental problems: (1) \textit{How to align the category feature between two distinct domains.} The traditional domain adaptation for node classification methods simply aligns the whole feature space of source and target domain, ignoring the specific category alignment, which would lead to feature confusion between different categories in target domain. Furthermore, due to the scarcity of target domain labels, how to align the specific categories' feature space is a challenge. (2) \textit{How to make use of the abundant unlabeled data in the target domain.} Previous domain adaptation relied on pseudo-labeling~\cite{zou2018unsupervised,liang2020we}, assigning labels to unlabeled data to supervise the neural network. However, this method can lead to overconfidence and noise in the labels, which can introduce biases and negatively impact performance. In the case of graph domains, the lack of labels can further exacerbate these issues. Thus, it is essential to devise a strategy that incorporates category-aligned features for effective node representation learning in the target domain.

In this paper, we address the aforementioned problems by formalizing a novel domain adaptive framework model named \textit{Spectral Augmentation for Graph Domain Adaptation (\method{})}. Firstly, we find that the source and target domains share the feature space of labels, nodes with the same category but belonging to different domains present similar characters in the spectral domain, while different categories are quite distinct. Due to the scarcity of target labels, directly aligning the category feature space is difficult. As an alternative, with the observation above, we design a spectral augmentation submodule, which combines the target domain spectral information with the source domain to enhance the node representation learning, instead of aligning the whole feature space in the spatial domain. Besides, we theoretical analysis the bound of \method{} to guarantee the stability of the model.
Then, considering that traditional GNN methods mainly focus on the aggregation of local information, ignoring the contribution of global information for node representation. We extract the local and global consistency of target domain to assist the node representation training. Lastly, we incorporate an adversarial scheme to effectively learn domain-invariant and semantic representations, thereby reducing the domain discrepancy for cross-domain node classification.
Through extensive experiments, we demonstrate that our \method{} significantly outperforms state-of-the-art competitors. Our contribution can be summed up as follows: 

\begin{itemize}[leftmargin=*]
    \item Under the graph domain adaptation settings, we have the observation that the spectral features of the same category in different domains present similar characteristics, while those of different categories are distinct. 
    % From the practical application demand of data analysis, we propose a domain adaptive framework \method{} for graph classification. As far as we know, we are the first to study the problem of unsupervised domain adaptation for graph classification.
    \item With the observation, we design the \method{} for node classification, which utilizes spectral augmentation to align the category features space instead of aligning the whole features space directly. Besides, we theoretical analyze the bound of \method{} to guarantee the stability of \method{}.
    % \method{} leverages adversarial learning to generate augmented views of source data for domain alignment in the graph space. Additionally, the representations at different layers of the network are explored to distill reliable pseudo-labels for better classification performance in the target domain. 
    \item Extensive experiments are conducted on real-world datasets, and the results show that our proposed \method{} outperforms the variety of state-of-the-art methods.
    % Extensive experiments on several datasets demonstrate the \method{} outperforms a variety of state-of-the-art competitors.
    
    \end{itemize}
% Then, we develop a dual graph convolutional network to jointly exploits local and global consistency for feature aggregation. Last, we use a domain classifier with an adversarial submodule to facilitate knowledge transfer between different domain graphs. 
% inspired by recent graph contrastive learning methods~\cite{you2020graph,you2021graph} that look into underlying priors using data augmentation strategies for self-supervised learning, we seek for meaningful perturbation directions on source graphs to reduce the domain discrepancy. An illustration of our learning process is shown in Figure \ref{fig:da}. 
% On the one hand, we train a domain discriminator conditioned on graph-level representations and semantic classifier predictions to holistically distinguish between different domains. On the other hand, we update the transformed views of source graphs to fuse the domain discriminator for effective domain alignment.
% Second, we generate accurate pseudo-labels for unlabeled samples to explore target semantics. Specifically, to mitigate the noise and biases in pseudo-labels resulting from the fixed graph encoder, we explore the semantics at different layers of the graph neural network, i.e., global semantics and local semantics. Then consistent pseudo-labels are distilled by comparing the clustering results of shallow representations to classifier predictions from deep representations. In this manner, we produce reliable pseudo-labels which enhance the classification performance on the target domain.

%% file: 3.related.tex
\section{Related Work}
\label{sec::related}

\subsection{Graph Neural Networks}
Graph neural networks (GNNs)~\cite{zhou2020graph} are designed to learn node embeddings in graph-structured data by mapping graph nodes with neighbor relationships to the low-dimensional latent space. Many approaches have been proposed to generalize the well-developed neural network models for regular grid structures, enabling them to utilize graph structured data for many downstream tasks, such as node classification and relationship prediction~\cite{wu2020comprehensive}, graph label noisy learning.
GCN~\cite{kipf2017semi} integrates node features as well as graph topology on graph-structured data and exploits two graph convolutional layers for node classification tasks by semi-supervised learning.
GAT~\cite{velickovic2017graph} enhances GCN in terms of message passing between nodes, using an attention mechanism to automatically integrate the features of the neighbors of a certain node.
However, most existing GNN-based approaches focus on learning the node representation among a particular graph. When transferring the learned model between different graphs to perform the same downstream task, representation space drift and embedding distribution discrepancy may be encountered.

\subsection{Unsupervised Domain Adaptation}
Unsupervised domain adaptation is one of the transfer learning methods that aims to minimize the discrepancy between the source and target domains, thus transferring knowledge from the well-labeled source domain to the unlabeled target domain\cite{farahani2021brief,sun2021focally}.
To perform cross-domain classification tasks, methods based on deep feature representation\cite{sun2016deep}, which map different domains into a common feature space, have attracted extensive attention. Most of these methods utilize adversarial training to reduce the inter-domain discrepancy\cite{long2015learning}. Typically, DANN\cite{ganin2016domain} utilizes a gradient reversal layer to capture domain invariant features, where the gradients are back-propagated from the domain classifier in a minimax game of the domain classifier and the feature extractor.

\begin{figure*}[t]
\centering
\includegraphics[width=14.5cm,keepaspectratio=true]{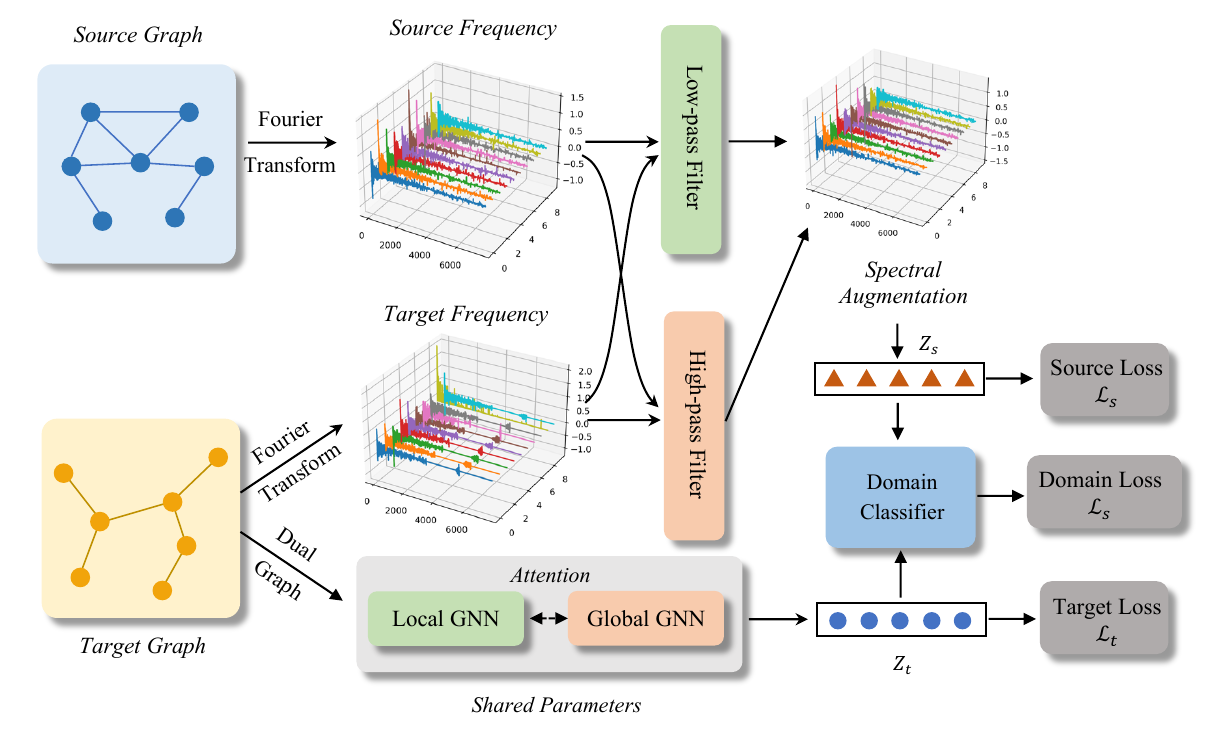}
\caption{Overview of our proposed \method{}. \method{} first applies spectral augmentation to align the category feature space. Then a dual graph module learns the consistency representation from local and global perspectives. Last, we introduce three classifiers to learn the semantic representation for training source, domain, and target classifier. }
\label{fig:framework}
% \vspace{-0.2cm}
\end{figure*}

Recently, for graph-structured data, several studies have been proposed for cross-graph knowledge transfer via unsupervised domain adaptation methods~\cite{shen2020adversarial,dai2019network,2020Unsupervised}.
CDNE\cite{shen2020adversarial} learns transferable node representations through minimizing the maximum mean discrepancy (MMD) loss for cross-network learning tasks, yet cannot model the network topology.
To enhance CDNE, AdaGCN\cite{dai2019network} utilizes graph convolutional networks for feature extraction to learn the node representations, and then to learn domain-invariant node representations based on adversarial learning.
UDA-GCN\cite{2020Unsupervised} further models both local and global consistency information for cross-graph domain adaptation. DEAL and CoCo are designed for the classification of graph-level domain adaptation.
However, most of the existing methods only align the feature space of the source and target domains, without considering the specific class alignment. Moreover, the unlabeled data are simply labeled based on pseudo-labeling. The above limitations may lead to the feature confusion among different categories in the target domain and the introduction of noise in the labels, which may negatively affect the model performance.

%% file: 4.method.tex
\section{Methodology}

\subsection{Problem Definition}
\textbf{Graph Node Classification:}
Given a graph $G = (\mathcal{V}, \mathcal{E}, \bm{A}, \bm{X}, \bm{Y})$ with the set of nodes $\mathcal{V}$ and the set of edges $\mathcal{E}$. $\bm{A}\in\mathbb{R}^{N\times N}$ is the adjacency matrix of $G$, $\bm{D}$ is the degree matrix with $\bm{D}_{ii}=\sum_{j=0}^N{\bm{A}_{ij}}$ and $N$ denotes the number of nodes. The node feature denotes as $\bm{X} \in \mathbb{R}^{N \times d}$, where each
row $\bm{x}_v \in \mathbb{R}^{d}$ represents the feature vector of node $v \in \mathcal{V}$ and $d$ is the dimension of node features.
The normalized graph Laplacian matrix is defined $\bm{L}=\bm{I}-\bm{D}^{-\frac{1}{2}}\bm{AD}^{-\frac{1}{2}}$, $\bm{I}$ is the identity matrix. By factorization of $\bm{L}$, we have $\bm{L}=\bm{U}\bm{\Lambda U}^\top$, where $\bm{U}=[\bm{u_1},\cdots,\bm{u_n}]$, $\bm{\Lambda}=diag([\lambda_1,\cdots,\lambda_n])$, $\bm{u_i}$ is the eigenvector and $\lambda_i$ is the corresponding eigenvalue.
$\bm{Y}\in\mathbb{R}^{N\times C}$ is the label of $G$, $C$ is the number of classes. 

\textbf{Graph Domain Adaptation for Node Classification:}
Given the fully labeled source graph $G_s=(\mathcal{V}_s, \mathcal{E}_s,\bm{A}_s,\bm{X}_s,\bm{Y}_s)$ and unlabeled target graph $G_t=(\mathcal{V}_t, \mathcal{E}_t,\bm{A}_t,\bm{X}_t)$, which shares the same label space $\mathcal{Y} =\{1,2,\cdots, C\}$ but distinct data distributions in the graph space, causing domain shifts. Our purpose is to train a classifier with the source $G_s$ and target $G_t$ for target node classification.
% In our settings, we are given a labeled source domain $\mathcal{D}^s = \{(G_i^s, y_i^s)\}_{i=1}^{N_s}$ containing $N_s$ examples with their associated labels and a unlabeled target domain $\mathcal{D}^t = \{G_j^t\}_{j=1}^{N_t}$ containing $N_t$ examples. $\mathcal{D}^s$ and $\mathcal{D}^t$ are obtained from the same label space $\mathcal{Y} =\{1,2,\cdots, C\}$ but have distinct data distributions in the graph space, which causes domain shifts.  local and global consistency 

\subsection{Overview}
The two key difficulties of unsupervised domain adaptation for node classification are the challenging category alignment of graph data and the inefficient node classification with label scarcity. To tackle these challenges, we propose a novel model termed \method{}. Our proposed \method{} consists of a spectral augmentation module, a dual graph learning module, and a domain adversarial module. The spectral augmentation module combines the spectral features between source and target domain to implement the category feature alignment. The dual graph module aims to learn the consistency representation from local and global perspectives, and the domain adversarial module to differentiate the source and target domains. 
% The framework is shown in Figure~\ref{fig:framework}.
% The proposed method mainly consists of the following three modules:
% \begin{itemize}[leftmargin=*]
%     \item \textbf{Spectral Augmentation for Category Alignment.} To resolve the category inconsistency, we first find the inherent characteristics of category on different domains in the spectral domain. Due to the scarcity of labels on the target domain, we apply spectral augmentation instead of aligning the category feature space directly.
%     \item \textbf{Attention-based Dual Graph Learning.} 
%     To resolve the domain inconsistency, we apply adaptive perturbations to source graphs which are trained against the domain classifier. Under the supervision of adversarial learning, the graph representations of source data can be used to enhance the target semantics learning, therefore mitigating the distribution discrepancy between two domains in the graph space. 
    
% \end{itemize}
\subsection{Spectral Augmentation for Category Alignment}
Domain adaptation methods in the past have typically used the pseudo-labeling mechanism~\cite{zou2018unsupervised,liang2020we} to align the feature space of categories. They involve assigning pseudo-labels to unlabeled data and using the labels to supervise the neural network. However, the risk of overconfidence and noise in these pseudo-labels would lead to potential biases during optimization and ultimately affect performance. Furthermore, in the graph domain, the lack of labels can worsen the noise issue.

\begin{figure}[t]
	\centering
	\begin{minipage}{0.49\linewidth}
		\centering
            \subfigure{
		\includegraphics[width=1\linewidth]{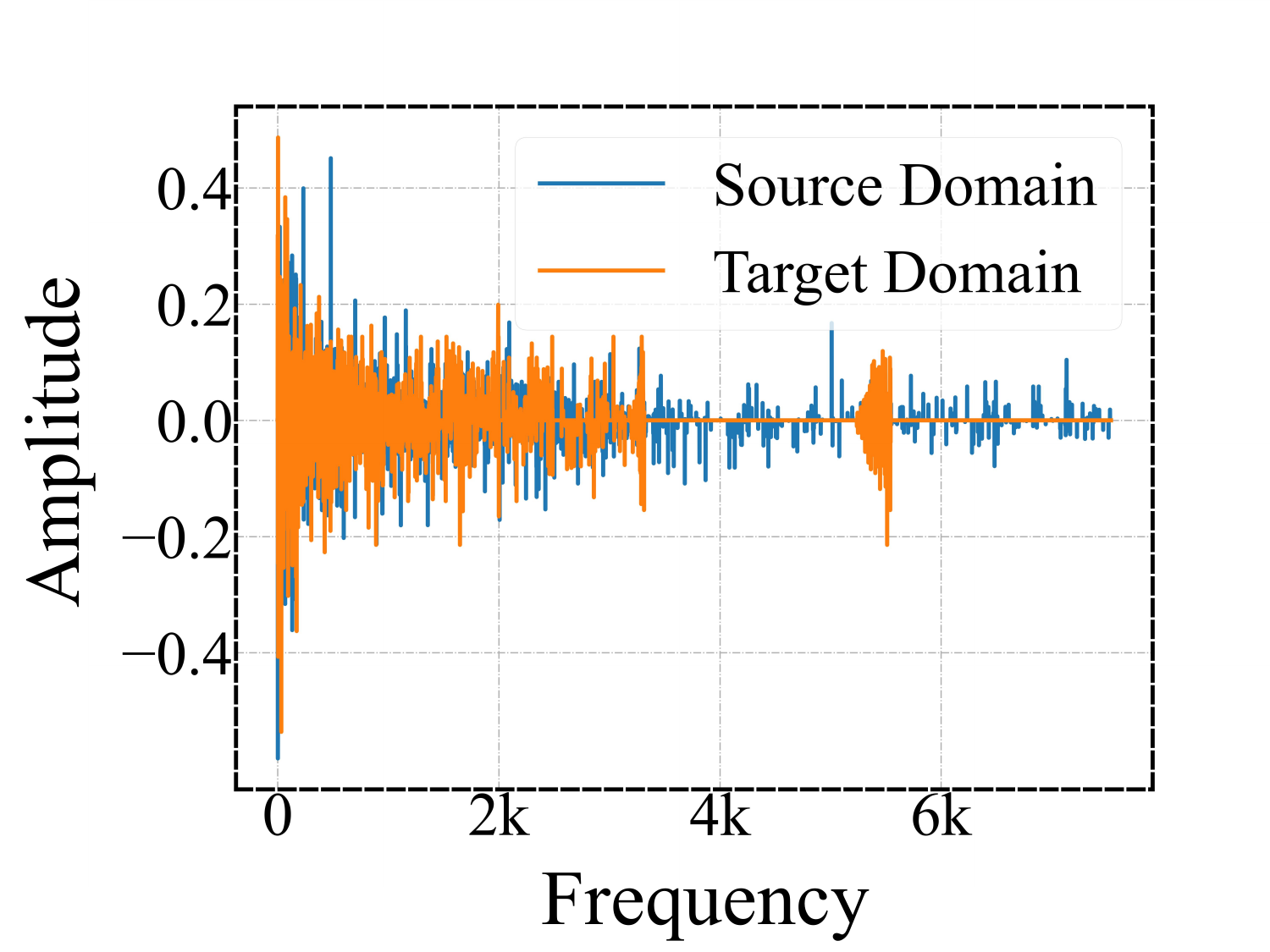}}
	\end{minipage}
	%\qquad
	\begin{minipage}{0.49\linewidth}
		\centering
            \subfigure{
		\includegraphics[width=1\linewidth]{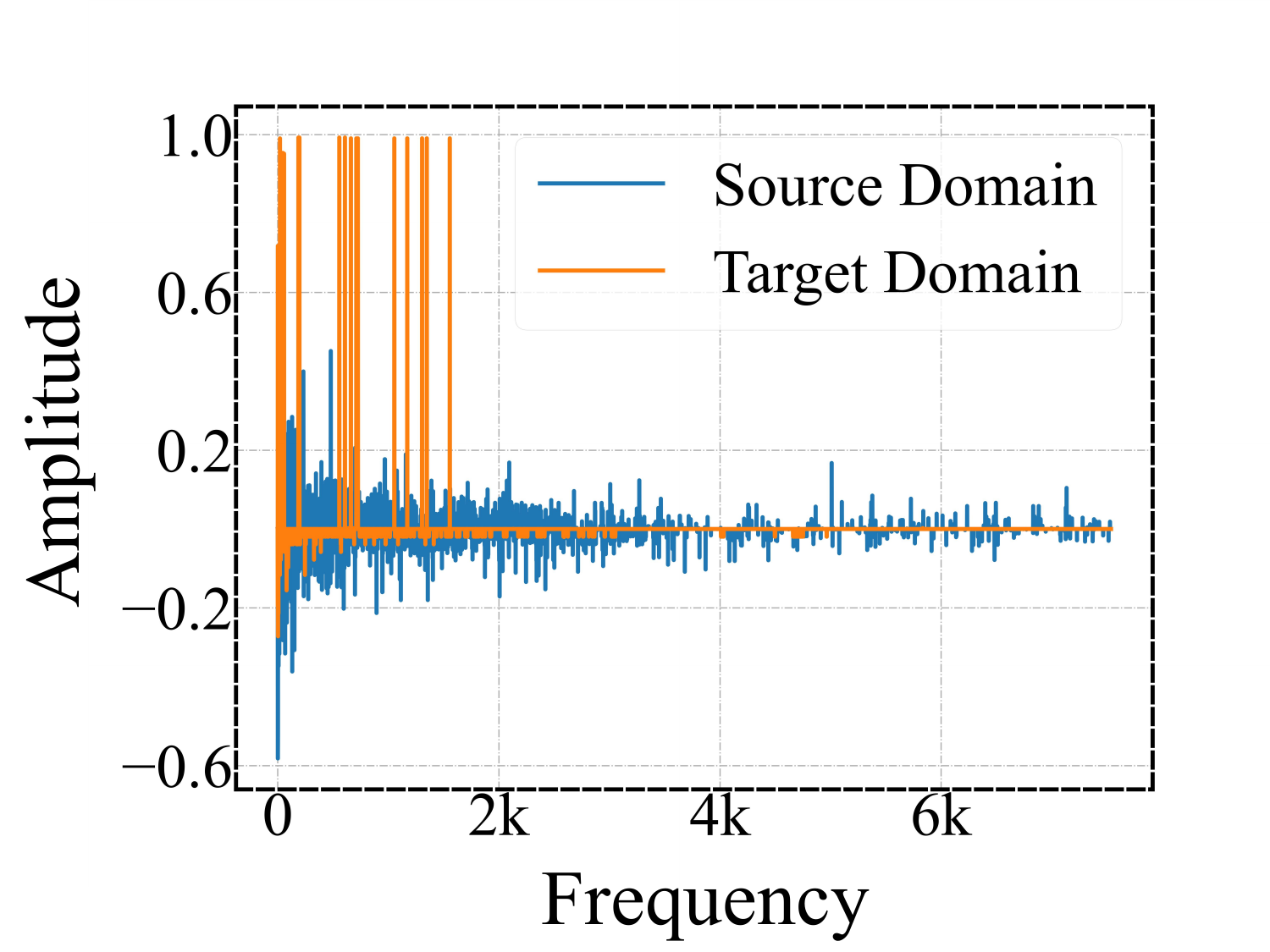}}
	\end{minipage}

        \caption{Left: The spectral of same category on different domains; Right: The spectral of different categories on different domains.}
    
    % \vspace{-0.4cm}
    \label{spectral}
\end{figure}

Aligning the category feature space in the spatial domain is difficult due to the above problem. Inspired by the transductive learning method~\cite{kipf2017semi}, we try to explore the characteristics of categories with different domains in the spectral domain. Extracting the spectral features of nodes with the same category in the source and target domains, we find that they show a high degree of correlation, while the spectral features of different categories are distinct, which is shown in Figure~\ref{spectral}.

With the observation, we design a spectral augmentation method for category alignment in the spectral domain instead of in the spatial domain.
From the perspective of signal processing~\cite{shuman2013emerging}, the graph Fourier is defined as $\bm{z}=\bm{U}^\top \bm{x}$, and the inverse Fourier transform is $\bm{x}=\bm{Uz}$. Then, the convolution on graph is:
$f\star \bm{X}=\bm{U}((\bm{U}^\top f)\odot(\bm{U}^\top \bm{X}))=\bm{U}g_\theta \bm{U}^\top \bm{X},$
where $f$ is the convolution kernel, and $\odot$ is the element-wise product operation. Considering that low-frequency would lose the discrimination of node representation, ~\cite{bo2021beyond} utilize the attention to combine the low and high-frequency for node representation. In our work, we first apply the low-pass filter $g_\theta^L(\bm{\Lambda}_d)$ and high-pass filter $g_d^H(\bm{\Lambda}_d)$ to separate the low and high-frequency signals, where $d$ can be $s$ or $t$ for source and target domain. Denote $\bm{z}^H_d$ and $\bm{z}^L_d$ as the high and low-frequency signals of node representation $z$ on domain $d$, then:
\begin{equation}
\bm{z}^H_d = \bm{U}_dg_\theta^H(\bm{\Lambda}_d)\bm{U}_d^\top \bm{x}, \quad
\bm{z}^L_d = \bm{U}_d g_\theta^L(\bm{\Lambda}_d)\bm{U}_d^\top \bm{x}.
\end{equation}

Instead of aligning the category feature on the spatial domain, we combine the spectral domain information of source and target domain directly, because of the inhere naturally aligned spectral of the same category. Formally, in the source domain:
\begin{equation}
\begin{aligned}
\label{source}
\bm{Z}_s=&\sigma(\bm{U}_s[\alpha g_\theta^H(\bm{\Lambda}_s)\bm{U}_s^\top \bm{X}_s\bm{W}+ (1-\alpha) g_\theta^H(\bm{\Lambda}_t)\bm{U}_t^\top \bm{X}_t\bm{W} \\
&+\beta g_\theta^L(\bm{\Lambda}_s)\bm{U}_s^\top \bm{X}_s\bm{W}+(1-\beta) g_\theta^L(\bm{\Lambda}_t)\bm{U}_t^\top \bm{X}_t\bm{W}]),
\end{aligned}
\end{equation}
where $\alpha$ and $\beta$ are the hyperparameters for fusing high and low frequencies between source and target domain. $\bm{U}_d$ and $\bm{\Lambda}_d$ denotes the eigenmatrix and eigenvalue of domain $d$, and $d\in\{s,t\}$, $\sigma$ is the activation function and $\bm{W}$ is the trainable matrix.

% \begin{figure*}[t]
% % \setlength{\abovecaptionskip}{-0.01cm} 
% \centering
% \begin{minipage}{4cm}
% \centering
% \subfigure{
% \includegraphics[width=1\textwidth]{IMG_NEW/Fig.31.pdf}}
% \centering
% \end{minipage}
% \begin{minipage}{4cm}
% \centering
% \subfigure{
% \includegraphics[width=1\textwidth]{IMG_NEW/Fig.31.plus.pdf}}
% \centering
% \end{minipage}
% \begin{minipage}{4cm}
% \centering
% \subfigure{
% \includegraphics[width=1\textwidth]{IMG_NEW/Fig.32.pdf}}
% \centering
% \end{minipage}
% \begin{minipage}{4cm}
% \centering
% \subfigure{
% \includegraphics[width=1\textwidth]{IMG_NEW/Fig.32.plus.pdf}}
% \centering
% \end{minipage}
% \caption{Left: The spectral of same category on different domains; Right: The spectral of different categories on different domains.}
% \label{spectral}
% \vspace{-0.1cm}
% \end{figure*}

%\begin{figure}[t]
%\centering
% \begin{minipage}{8cm}
%\centering
% \subfigure[Signal fusion radio $r$]{
%\includegraphics[width=9cm,keepaspectratio=true]{IMG_NEW/Fig.3.pdf}
%\includegraphics[width=0.45\textwidth]{IMG_NEW/sig_compare_1.pdf}
% \centering
% \end{minipage}
% \begin{minipage}{8cm}
% \centering
% \hspace{0.3cm}
% \subfigure[Embedding feature dimensions]{
%\includegraphics[width=0.45\textwidth]{IMG_NEW/sig_compare_2.pdf}
%\centering
% \end{minipage}
%\caption{Left: The spectral of same category on different domains; Right: The spectral of different categories on different domains.}
%\vspace{-0.2cm}
%\label{spectral}
%\end{figure}

\subsection{Attention-based Dual Graph Learning}
To capture the local and global information of nodes, we propose a dual graph neural network, where the local consistency is extracted with the graph adjacency matrix and the global with the random walk GNN. Due to the source domain graph being fused with the target information from the global view, we apply dual graph learning on the target domain.

\textbf{Local GNN.}
For the local consistency extraction, we use the GCN~\cite{graphsage2017} method directly. With the input $\bm{X}_t$ and adjacency matrix $\bm{A}_t$, the local node representation is defined as:
\begin{equation}
\label{local}
\bm{Z}_{t,local}^{l}=\sigma(\tilde{\bm{D}}^{-\frac{1}{2}} \Tilde{\bm{A}} \tilde{\bm{D}}^{-\frac{1}{2}} \bm{Z}_t^{l-1} \bm{W}^l),
\end{equation}
where $\tilde{\bm{A}}=\bm{I}+\bm{A}_t$ is the adjacent matrix with self-loop, $\tilde{\bm{D}}_{ii}=\sum_j \tilde{\bm{A}}_{ij}$, $\bm{W}^l$ is the trainable matrix on the $l$-th layer. $\bm{Z}_t^l$ is the node embeddings of target domain on the $l$-th layer and $\bm{Z}_t^0=\bm{X}_t$.

\textbf{Global GNN.}
To achieve global information, similar to~\cite{2020Unsupervised}, we apply the random walk to calculate the semantic similarities between nodes and calculate the global consistency with the newly constructed adjacency matrix.

Define the state of node $v_i$ at time $t$ as $s(t)=v_i$. The transition probability of jumping $v_i$ to the neighbors $v_j$ is defined as:
\begin{equation}
    \bm{P}_{ij}=p(s(t+1)=v_j|s(t)=v_i)=\frac{\bm{A}_{t(ij)}}{\sum_j \bm{A}_{t(ij)}},
\end{equation}
 where $\bm{A}_{t(ij)}$ denotes the value of $\bm{A}_{ij}$ on the target domain. Then, we calculate the point-wise mutual information matrix~\cite{klabunde2002daniel} as: 
\begin{equation}
\bm{M}_{ij}=max\{log(\frac{m_{ij}}{m_{i\star} m_{\star j}}),0\},
\end{equation}
where $m_{ij}=\frac{\bm{P}_{ij}}{\sum_{i,j}\bm{P}_{ij}}$, $m_{i\star}=\frac{\sum_j \bm{P}_{ij}}{\sum_{i,j}\bm{P}_{ij}}$ and $m_{\star j}=\frac{\sum_i \bm{P}_{ij}}{\sum_{i,j}\bm{P}_{ij}}$.

With the calculated adjacency matrix $\bm{M}$, we extract the global node information as follows:
\begin{equation}
\bm{Z}_{t,global}^{l}=\sigma(\bm{D}^{-\frac{1}{2}} \bm{M} \bm{D}^{-\frac{1}{2}} \bm{Z}_t^{l-1} \bm{W}^l),
\end{equation}
where $\bm{D}_{ii}=\sum_j \bm{M}_{ij}$, and $\bm{W}^l$ is the shared parameters with Eq.~\ref{local}.

\textbf{Attention-based fusion.} To fuse the local and global information, we utilize the attention mechanism $a:\mathbb{R}^{d'}\times \mathbb{R}^{d'} \to \mathbb{R}$ to computer the attention coefficients:
\begin{equation}
\label{target}
\begin{aligned}
e_{ij}&=a(\bm{Wz}_{i,local},\bm{Wz}_{j,global}),\\
\zeta_{ij}=&softmax(e_{ij})=\frac{exp(e_{ij})}{\sum_k exp(e_{ij})},\nonumber
\end{aligned}
\end{equation}
where $\bm{W} \in \mathbb{R}^{d'\times d}$ is the shared matrix, and we choose $a(\bm{h}_i,\bm{h}_j)=LeakyReLu(\bm{W}_0^\top[\bm{Wh}_i||\bm{Wh}_j])$ with learnable matrix $\bm{W}_0\in \mathbb{R}^{2d'}$ and $||$ denotes concatenation operation.
The fused representation of local and global is calculated as:
\begin{equation}
    \bm{Z}_t=\zeta_{ii}\bm{Z}_{t,local} + (1-\zeta_{ii})\bm{Z}_{t,global}.
\end{equation}

\subsection{Domain Adversarial Training}

The spectral augmentation of the category alignment module enforces the similarity of features across different domains, confusing source and target domain features. Our goal is to maximize the domain classification error while minimizing the source domain classification error, i.e.,
\begin{equation}
    \mathop{min}\limits_{\theta_y} \mathop{max}\limits_{\theta_d}=\mathcal{L}_s(f(\bm{z}_s;\theta_y))-\gamma \mathcal{L}_D(h(\bm{z};\theta_d)),
\end{equation}
where $f(\bm{z}_s;\theta_y)=softmax(\theta_y \bm{z}_s)$ is the source classification function, and $\theta_y$ is the trainable parameters for $f$.
$h(\bm{z};\theta_d)=sigmod(\theta_d\bm{z})$ is the domain classifier, $\theta_d$ is the parameters for $h$. $\mathcal{L}_s$ and $\mathcal{L}_D$ are the source classification loss and domain classification loss. $\bm{z}_s$ denotes the source features and $\bm{z}$ denotes the features of the source and target domains.

% The spectral augmentation for category alignment module enforces the features of different domains to be similar and obfuscates source and target domain data. making it difficult to distinguish whether one node comes from $G_s$ or $G_t$. 
% To solve this, we apply the Gradient Reversal Layer (GRL)~\cite{ganin2016domain} for domain adversarial training.

% \begin{equation}
%     \mathcal{L}_D=-\frac{1}{N_s+N_t}\sum_{i=1}^{N_s+N_t}d_i log(\hat{d}_i)+(1-d_i)log(1-\hat{d}_i),
% \end{equation}
% where $d_i\in \{0,1\}$ denotes node $i$ belongs to the source or target domain, and $\hat{d}_i$ is the domain prediction.

% Similar to~\cite{ganin2016domain}, 
% \begin{align}\label{eq:per_1}
%  \min_{||\delta(\cdot)||_F\leq \epsilon} \max_{\theta_{d}} 
% \mathcal{L}_{D}= & \mathbb{E}_{G^s_i \in \mathcal{D}^s} \log D(F(G^s_i; \bm{M}^s_i + \bm{\delta}(\bm{M}^s_i)), \hat{\bm{p}}^s_i)  \nonumber\\ + &  \mathbb{E}_{G^t_j\in \mathcal{D}^t} \log (1-D(F(G^t_j), \hat{\bm{p}}^t_j)), \end{align}
% where $\theta_{d}$ is the parameters of the domain discriminator and $\epsilon$ is the maximum of perturbation size, equivalent to the learning rate. $\bm{M}^s_i$ can be $\bm{X}^s_i$ or ${\bm{H}}^{(1),s}_i$ for the first or second strategy, respectively. We concatenate the graph representation and classifier prediction and feed it to the domain discriminator implemented by an MLP.  

\subsection{Optimization}

Our \method{} has the overarching goal that combines the domain adversarial loss and classification loss of the target data. Additionally, minimizing the expected source error for labeled source samples is also essential. In formation, 

\begin{equation}\label{loss}
\begin{aligned}
    \mathop{min}\limits_{\theta_y} \mathop{max}\limits_{\theta_d}\mathcal{L}(\theta_y,\theta_d) =& \mathcal{L}_{s}(f(\bm{z}_s;\theta_y)) + \gamma_1\mathcal{L}_{t}(f(\bm{z}_t;\theta_y))\\& - \gamma_2 \mathcal{L}_{D}((h(\bm{z};\theta_d)),
\end{aligned}
\end{equation}
with the source loss $\mathcal{L}_s$, target loss $\mathcal{L}_t$ and domain loss $\mathcal{L}_D$:
\begin{equation}
\begin{aligned}
\mathcal{L}_s=-&\frac{1}{N_s}\sum_{i=1}^{N_s}y_ilog(f(\bm{z}_s)),\\
\mathcal{L}_t=-&\frac{1}{N_t}\sum_{i=1}^{N_t}f(\bm{z}_t)log(f(\bm{z}_t)),\\
\mathcal{L}_D=-\frac{1}{N_s+N_t}&\sum_{i=1}^{N_s+N_t}d_i log\frac{1}{h(\bm{z})}+(1-d_i)log\frac{1}{1-h(\bm{z})},\nonumber
\end{aligned}
\end{equation}
where $\gamma_1$ and $\gamma_2$ are hyper-parameters to balance the domain adversarial loss and target classification loss. $d_i\in \{0,1\}$ denotes node $i$ belongs to the source or target domain. To update the parameters of Eq.~\ref{loss} in the standard stochastic gradient descent (SGD) method, we apply the Gradient Reversal Layer~\cite{ganin2016domain} (GRL) $\mathcal{R}$ for model training, which is formulate as:
\begin{equation}
    \mathcal{R}(\bm{x})=\bm{x},\quad \frac{d\mathcal{R}}{d\bm{x}}=-\bm{I}.
\end{equation}
Then, the update of Eq.~\ref{loss} can be implemented with SGD with the following formulation:
\begin{equation}
\label{total_loss}
\begin{aligned}
    \mathop{min}\limits_{\theta_y,\theta_d}\mathcal{L}(\theta_y,\theta_d) =& \mathcal{L}_{s}(f(\mathcal{R}(\bm{z}_s);\theta_y)) + \gamma_1\mathcal{L}_{t}(f(\mathcal{R}(\bm{z}_t);\theta_y))\\& - \gamma_2 \mathcal{L}_{D}((h(\mathcal{R}(\bm{z});\theta_d)).
\end{aligned}
\end{equation}
The whole learning procedure is shown in Algorithm~\ref{alg1}.

% Eq. \ref{eq:final} is minimized by the mini-batch standard stochastic gradient descent (SGD) method.
%  The whole learning procedure is shown in Algorithm \ref{alg1}.

% \begin{equation}
% \mathcal{L}_{S}=\mathbb{E}_{G_i^s \in \mathcal{D}^s} \mathcal{E} (H(\hat{\bm{z}}_i^s), y_i^s),
% \end{equation}
% where $\hat{\bm{z}}_i^s$ denotes the representations of source data.
% Consequently, to update the network parameters, i.e., $\{\theta_e, \theta_c, \theta_{d}\}$, we minimize the ultimate objective as follows:

% \begin{algorithm}[t]
% \caption{Training Algorithm of \method{}}
% \label{alg1}
% \begin{algorithmic}[1]
% \REQUIRE Source domain adjacency matrix $A_s$, degree matrix $D_s$; target domain adjacency matrix $A_t$, degree matrix $D_t$, updating steps $E$; \
% \ENSURE Parameters $\theta$ for the neural network;
% \STATE Initialize $\theta$;
% % \STATE $\mathcal{Q} \leftarrow \emptyset$ ;
% % \REPEAT
% % \REPEAT
% \STATE Transform the source and target spatial graph features into spectral domain.
% \FOR{$e=1,2, \cdots, E$}
% \STATE Calculate the spectral augmentation features on the source domain $Z_s$ with Eq.~\ref{source};
% \STATE Calculate the local and global features with dual graph on the target domain $Z_t$ with Eq.~\ref{target};
% \STATE Update the network parameters $\theta$ through backpropagation by Eq. \ref{total_loss};
% \ENDFOR

% % \UNTIL convergence
% % \STATE Generate each $l_j$;
% % \STATE Cluster the shallow features to get each $e_j$;
% % \STATE Updating $\mathcal{Q}$ by Eq. \ref{eq:Q};
% % \UNTIL convergence

% \end{algorithmic}
% \end{algorithm}

% \vspace{-1cm}

\begin{algorithm}[t]
\caption{Training Algorithm of \method{}}
\label{alg1}
\begin{algorithmic}[1]
\REQUIRE Source domain adjacency matrix $A_s$, degree matrix $D_s$; target domain adjacency matrix $A_t$, degree matrix $D_t$, updating steps $E$; \
\ENSURE Parameters $\theta$ for the neural network;
\STATE Initialize $\theta$;
% \STATE $\mathcal{Q} \leftarrow \emptyset$ ;
% \REPEAT
% \REPEAT
\STATE Transform the source and target spatial graph features into spectral domain.
\FOR{$e=1,2, \cdots, E$}
\STATE Calculate the spectral augmentation features on the source domain $Z_s$ with Eq.~\ref{source};
\STATE Calculate the local and global features with dual graph on the target domain $Z_t$ with Eq.~\ref{target};
\STATE Update the network parameters $\theta$ through backpropagation by Eq. \ref{total_loss};
\ENDFOR

% \UNTIL convergence
% \STATE Generate each $l_j$;
% \STATE Cluster the shallow features to get each $e_j$;
% \STATE Updating $\mathcal{Q}$ by Eq. \ref{eq:Q};
% \UNTIL convergence
\end{algorithmic}
\end{algorithm}

\subsection{Theoretical Analysis}
In this subsection, we proof the stability of \method{}, which is inspired by~\cite{you2023graph}:
\begin{lemma}
Suppose the $G_s$ and $G_t$ are the graphs of the source and target domain. Given the Laplace matrices decomposition $\bm{L}_d=\bm{D}_d-\bm{A}_d=\bm{U}_d\bm{\Lambda}_d\bm{U}_d^\top$, where $\bm{\Lambda}_d=diag[\lambda_{d1},\cdots,\lambda_{dn}]$ are the sorted eigenvalues of $L_d$, and $d\in\{s,d\}$ denotes the source and target domains. The GNN is constructed as $f(G)=\sigma((g_\theta(\bm{L})\bm{XW})=\sigma(\bm{U}g_\theta(\bm{\Lambda})\bm{U}^\top \bm{XW})$, where $g_\theta$ is the polynomial function with $g_\theta(\bm{L})=\sum_{k=0}^{K}\theta_k \bm{L}^k$, $\bm{W}$ is the learnable matrix and the pointwise nonlinearity has $|\sigma(b)-\sigma(a)|\le |b-a|$. Assuming $||\bm{X}||_{op} \le 1$ and $||\bm{W}||_{op} \le 1$, we have the following inequality:
\begin{equation}
\begin{aligned}
||f(G_s+G_t)-f(G_t)||_2
\le &\alpha [C_\lambda (1+\tau)||\bm{L}_s -\bm{P}^\star \bm{L}_t \bm{P}^{\star \top}||_F\\
&+\mathcal{O}(||\bm{L}_s-\bm{P}^\star \bm{L}_t \bm{P}^{\star \top}||_F^2) \\
&+ max(|g_\theta(\bm{L}_t)|)||\bm{X}_s-\bm{P}^\star \bm{X}_t||_F ],
\end{aligned}
\end{equation}
where $\tau=(||\bm{U}_s-\bm{U}_t||_F+1)^2-1$ stands for the eigenvector misalignment which can be bounded. $\Pi$ is the set of permutation matrices, and $\bm{P}^\star=argmin_{\bm{P}\in\Pi}{||\bm{X}_s-\bm{PX}_t||_F+||\bm{A}_s- \bm{PA}_t\bm{P}^\top||_F}$. $\mathcal{O}(||\bm{L}_s-\bm{P}^\star \bm{L}_t \bm{P}^{\star \top}||_F^2$ is the remainder term with bounded multipliers defined in~\cite{gama2020stability}, and $C_\lambda$ is the spectral Lipschitz constant that $\forall \lambda_i, \lambda_j, |g_\theta (\lambda_i)-g\theta (\lambda_j)|\le C_\lambda (\lambda_i -\lambda_j)$.

\label{prop}
\end{lemma}

%% file: 5.experiment.tex
\begin{table}[t]
\caption{The specific statistics of the experimental datasets.}%标题
\centering%把表居中
\begin{tabular}{ccccc}%四个c代表该表一共四列，内容全部居中
\toprule%第一道横线
Dataset& \#Nodes&\#Edges&\#Features&\#Labels \\
\midrule%第二道横线
DBLPv7&5484&8130&6775&6 \\
ACMv9&9360&15602&6775&6 \\
Citationv1&8935&15113&6775&6 \\
\bottomrule%第三道横线
\end{tabular}
\label{dataset}
% \vspace{-0.2cm}
\end{table}

\section{Experiment}
\label{sec::experiment}
In this section, we conduct extensive experiments on various real-world datasets to verify the effectiveness of the proposed \method{}. We aim to answer the questions below:
\begin{itemize}
    \item \textbf{RQ1}: How does the proposed \method{} perform compared with the state-of-the-art baseline methods for node classification?
    % \item \textbf{RQ2}: How do different GNN-based encoders affect the performance of the proposed \method{}?
    \item \textbf{RQ2}: How is the effectiveness of proposed components on the performance?
    \item \textbf{RQ3}: How does the hyper-parameters affect the performance of the proposed \method{}?
    \item \textbf{RQ4}: How about the intuitive effect of proposed \method{}? 
\end{itemize}

% \begin{table}[t]
% \caption{The specific statistics of the experimental datasets.}%标题
% \centering%把表居中
% \begin{tabular}{ccccc}%四个c代表该表一共四列，内容全部居中
% \toprule%第一道横线
% Dataset& \#Nodes&\#Edges&\#Features&\#Labels \\
% \midrule%第二道横线
% DBLPv7&5484&8130&6775&6 \\
% ACMv9&9360&15602&6775&6 \\
% Citationv1&8935&15113&6775&6 \\
% \bottomrule%第三道横线
% \end{tabular}
% \label{dataset}
% % \vspace{-0.2cm}
% \end{table}

\begin{table*}[t]
\caption{Classification accuracy comparisons on six cross-domain tasks.}%标题
\centering%把表居中
\setlength{\tabcolsep}{5mm}{
\begin{tabular}{cccccccc}%四个c代表该表一共四列，内容全部居中
\toprule%第一道横线
Methods& C$\to$D& A$\to$D&	D$\to$C&	A$\to$C&	D$\to$A&	C$\to$A&	average \\
\midrule%第二道横线
DeepWalk\cite{0DeepWalk}&	0.1222& 	0.2308& 	0.2210& 	0.2660& 	0.1437& 	0.2144& 	0.1997 \\ 
LINE\cite{2015LINE}&	0.1938& 	0.2604& 	0.1976& 	0.3317& 	0.1576& 	0.2027& 	0.2240 \\ 
GraphSAGE\cite{2017Inductive}&	0.5487& 	0.4443& 	0.4860& 	0.5043& 	0.5749&	0.4794& 	0.5063 \\
DNN\cite{montavon2018methods}&	0.3699&	0.3656&	0.3913&	0.4407&	0.3363&	0.4059&	0.3850 \\
GCN\cite{kipf2017semi}&	0.5467&	0.4392&	0.4871&	0.5039&	0.5184&	0.4785&	0.4956 \\
DGC\cite{wang2021dissecting}&	0.5587& 	0.4589& 	0.5178& 	0.5250& 	0.5298& 	0.4890& 	0.5132 \\ 
SUBLIME\cite{2022Towards}&	0.5650& 	0.4699& 	0.5012& 	0.5392& 	0.5247& 	0.5220& 	0.5203 \\ 
\midrule%第二道横线
DGRL\cite{2016Domain}&	0.3699&	0.3756&	0.3905&	0.4514&	0.3439&	0.4063&	0.3896 \\
AdaGCN\cite{sun2021adagcn}&	0.5516&	0.4470&	0.4872&	0.5094&	0.5752&	0.4884&	0.5098 \\ 
UDA-GCN\cite{2020Unsupervised}&	0.6599&	0.4710& 	0.5809&	0.5229&	0.5959& 	0.5498&	0.5634 \\ 
\midrule
\method{}(ours)&	\textbf{0.7097}&	\textbf{0.6444}& 	\textbf{0.6460}&	\textbf{0.6420}&	\textbf{0.6296}& 	\textbf{0.6136}&	\textbf{0.6476} \\
\bottomrule%第三道横线
\end{tabular}}
\label{results}
% \vspace{0.4cm}
\end{table*}

\subsection{Benchmark Datasets}
To evaluate the effectiveness of the proposed \method{}, we have developed three citation networks from ArnetMiner\cite{2008ArnetMiner}, including DBLPv7, ACMv9, and Citationv1. These datasets are derived from different data sources (DBLP, ACM and Microsoft Academic Graph) and  distinct time periods. Each sample contains a title, an index, a category and a citation index, where the category represents the academic field, including "Artificial intelligence", "Computer vision", "Database", "Data mining", "High Performance Computing" and "Information Security". Following \cite{0Attraction}, these datasets are constructed as undirected citation networks, where a node represents a paper, an edge indicates a citation record, and a label indicates the paper category. Since these three graphs are generated from different data sources and time periods, their distributions  are naturally diverse. Thus, we can conduct six cross-domain node classification tasks (source$\to$target), including C$\to$D, A$\to$D, D$\to$C, A$\to$C, D$\to$A, and C$\to$A, where D, A, C denote DBLPv7, ACMv9, and Citationv1, respectively. The statics of these datasets is shown in Table~\ref{dataset}.

\subsection{Baselines} %加引用
To verify the effectiveness of our method, we select the following methods as baselines for comparison, including seven state-of-the-art single-domain node classification methods (i.e., DeepWalk\cite{0DeepWalk}, LINE\cite{2015LINE}, GraphSAGE\cite{2017Inductive}, {DNN\cite{montavon2018methods}}, GCN\cite{kipf2017semi}, DGC\cite{wang2021dissecting}, and \\SUBLIME\cite{2022Towards}), and three cross-domain classification methods with domain adaptation (i.e., DGRL\cite{2016Domain}, AdaGCN\cite{sun2021adagcn} and UDA-GCN\cite{2020Unsupervised}). 
% The details of baselines are introduced in Appendix~\ref{baseline}.

\textbf{Single-domain node classification methods: }
\begin{itemize}[leftmargin=*]
    \item DeepWalk\cite{0DeepWalk}: DeepWalk employs the random walk sampling strategy to capture the neighborhood node structure. Then, following Skip-Gram\cite{guthrie2006closer}, which aims to learn the low-dimensional node representation on single domain.
    % , which is a classic method for single domain network representation.
    \item LINE\cite{2015LINE}: LINE is a classic method for large-scale graph representation learning, which preserves both first and second-order proximities for undirected network to measure the relationships between two nodes. Compared with the deep model, LINE has a limited representation ability.
    \item GraphSAGE\cite{2017Inductive}: GraphSAGE learns the node representation by aggregating the sampled neighbors for final prediction.
    \item DNN\cite{montavon2018methods}: DNN is a multi-layer perceptron (MLP) based method, which only leverages node features for node classification.
    \item GCN\cite{kipf2017semi}: GCN is a deep convolutional network on graphs, which employs the symmetric-normalized aggregation method to  learn the embedding for each node.
    \item DGC\cite{wang2021dissecting}: DGC is a linear variant of GCN, which separates the feature propagation steps from the terminal time, enhancing flexibility and enabling it to utilize a vast range of feature propagation steps.
    \item SUBLIME\cite{2022Towards}: SUBLIME generates an anchor graph from the raw data and uses the contrastive loss to optimize the consistency between the anchor graph and the learning graph, thus tackling the unsupervised graph structure learning problem.

\end{itemize}

% For the above methods, we train a classifier by learning the node representations of the source domain, and then we use the classifier to make classification predictions for nodes in the target domain.

\textbf{Cross-domain node classification methods}:
\begin{itemize}[leftmargin=*]
    \item DGRL\cite{2016Domain}: DGRL employs a 2-layer perceptron to extra feature and a gradient reverse layer (GRL) to learn node embedding for domain classification. 
    \item AdaGCN\cite{sun2021adagcn}: AdaGCN uses a GCN as the feature extractor and a gradient reverse layer(GRL) to train a domain classifier.
    \item UDA-GCN\cite{2020Unsupervised}: UDA-GCN utilizes a dual graph convolutional network to preserve both local and global consistency information for generating better node embedding features through the attention mechanism. Also, a gradient reverse layer(GRL) is added to train a domain classifier.
\end{itemize}

% These are the current state-of-the-art methods for graph domain adaption that allow both the source and target domains to participate in the training process of the node classifier.

\begin{table*}
\caption{Classification accuracy comparisons between \method{} variants on six cross-domain tasks.}%标题
\centering%把表居中
\setlength{\tabcolsep}{5mm}{
\begin{tabular}{cccccccc}%四个c代表该表一共四列，内容全部居中
\toprule%第一道横线
Methods& C$\to$D& A$\to$D&	D$\to$C&	A$\to$C&	D$\to$A&	C$\to$A&	average \\
\midrule%第二道横线
\method{}$\neg l$&	0.3307& 	0.3298& 0.3047& 	0.2612& 	0.2958& 	0.2694& 	0.2986 \\
\method{}$\neg h$&	0.6685& 	0.5746& 	0.5795& 	0.5449& 	0.5620& 	0.5345& 	0.5773  \\ 
\method{}$\neg p$&	0.5694&	    0.4659&	0.5071&	0.5259&	0.5401&	0.4998&	0.5180 \\
\method{}$\neg d$&	0.6982&	    0.5997&	0.6321&	0.5747&	0.5259& 	0.5567&	0.5979 \\
\method{}$\neg t$&	0.6657& 	0.5915&	0.5871& 	0.5264& 	0.5601& 	0.5289& 	0.5766 \\
\midrule
\method{}&	\textbf{0.7097}&	\textbf{0.6444}& 	\textbf{0.6460}&	\textbf{0.6420}&	\textbf{0.6296}& 	\textbf{0.6136}&	\textbf{0.6476} \\
\bottomrule%第三道横线
\end{tabular}}
\label{ablation}
\end{table*}

\subsection{Experimental Settings}
We utilize PyTorch~\cite{paszke2017automatic} and PyTorch Geometric library~\cite{Fey/Lenssen/2019} as the deep-learning Framework and Adam~\cite{KingmaB14} as an optimizer. We follow the principles of evaluation protocols used in unsupervised domain adaptation to perform a grid study of all methods in the hyperparameter space and show the best results obtained for each method. To be fair, for all the cross-domain node classification methods in this experiment, we use the same parameter settings, except for some special cases. For each method, we set the learning rate to $1e^{-4}$.
For the specific {GCN-based} models (e.g., GCN, DGC, SUBLIME, AdaGCN, and UDA-GCN), the number of hidden layers is set to 2, and the hidden dimensions are set to 128 and 16 regardless of the source and target domains. We use the same settings as above for both the local and global GNN.
For DeepWalk and LINE, we first learn node embeddings and then train node classifiers using information from source domain, the hidden dimension of the node embeddings is set to 128. {For DNN and DGRL, we set the same hidden dimensions as GNN-based models.}
% The adaptation rate $\lambda$ is set as indicated below:
% $\lambda=\min \left ( \frac{2}{1+\exp \left ( -10p \right ) }-1, 0.1 \right )$
% , where $p$ varies from 0 to 1 throughout the training process as in \cite{2016Domain}. 
The balance parameters $\gamma_{1}$, $\gamma_{2}$ are set to 0.3 and 0.1, and the dropout rate for each layer of dual GNN to 0.3. For simplicity, we set the combination weight of high and low-frequency as the same, i.e., $\alpha=\beta$.

\subsection{Performance Comparison (RQ1)}%改格式

We present the performance of \method{} compared with baselines under the setting of graph domain adaptation for node classification to answer \textbf{RQ1}, which is shown in Table~\ref{results}. From the results:

% To demonstrate the performance of the baseline model in comparison with our proposed model, we perform six cross-domain node classification tasks separately using the above method and record the results in Table 2. Based on these results, we obtained the following findings.

\begin{itemize}[leftmargin=*]
    \item DeepWalk and LINE achieve the worst performance among all the methods, {this is because they only utilize the network topology to generate node embedding without exploiting the node feature information.} DNN also achieves a poor performance, contrary to the above two methods, it only considers the node features and ignores the association between nodes, so it cannot generate better node embeddings.
    \item The graph-based methods (GCN, GraphSAGE, DGC, and SUBLIME) outperform DeepWalk and LINE, we attribute the reason to that they encode both local graph structure and node features to obtain better node embeddings. 
    \item DGRL, AdaGCN, and UDA-GCN achieve better performance than the single-domain node classification methods. The reason is that, by incorporating the domain adversarial learning, the node classifier is capable to transfer the knowledge from the source to the target domain.
    % It is worth mentioning that UDA-GCN performs the best among these three methods for all six tasks by simultaneously encoding the local as well as global consistency relations for each graph and training the three classifiers (source, domain, and target classifier) concurrently to reduce the discrepancy between domains.
    \item The proposed \method{} model achieves the best performance in these six cross-domain node classification datasets compared to all the baseline methods. By fusing the low and high-frequency signals of different domains in the spectral domain, \method{} implements the category alignment under the domain-variant setting. This enables us to learn better graph node embeddings and to train node classifiers in both the source and target domains through an adversarial approach, which greatly reduces the distribution discrepancies between the different domains and improves the node classifier performance.
\end{itemize}

\subsection{Ablation Study (RQ2)}

To answer \textbf{RQ2}, we introduce several variants of the \method{} to investigate the effectiveness of each component of \method{}: 
\begin{itemize}[leftmargin=*]
    \item \textbf{\method{}$\neg l$}, which evaluates the impact of low-frequency signals in both source and target domains.
    \item \textbf{\method{}$\neg h$}, evaluating the influence of high-frequency signals in different domains.
    \item  \textbf{\method{}$\neg p$}, which remove the global GNN and utilizing only the local GNN.
    \item \textbf{\method{}$\neg d$}, which removes the domain loss to evaluate the impact of domain adversarial learning.
    \item \textbf{\method{}$\neg t$}, which evaluate the effectiveness of target classifier by removing the target loss.
\end{itemize}
% (1) \textbf{\method{}$\neg l$}, which evaluates the impact of low-frequency signals in both source and target domains; (2) \textbf{\method{}$\neg h$}, evaluating the influence of high-frequency signals in different domains; (3) \textbf{\method{}$\neg p$}, which remove the global GNN and utilizing only the local GNN; (4) \textbf{\method{}$\neg d$}, which removes the domain loss to evaluate the impact of domain adversarial learning, and (5) \textbf{\method{}$\neg t$}, which evaluate the effectiveness of target classifier by removing the target loss. 
The results are reported in Table~\ref{ablation}, and we have the following observation:
% The following variants of the \method{} are all designed for comparison purposes.

% \begin{itemize}[leftmargin=*]
% \item \method{}$\neg l$: a variant of \method{} with low-frequency signals removed and only high-frequency signals considered when performing feature fusion in the frequency domain.
% \item \method{}$\neg h$: a variant of \method{} that removes high-frequency signals and performs feature fusion in the frequency domain using only low-frequency signals.
% \item \method{}$\neg p$: a variant of \method{} by removing the GNN channel that captures global consistency information ($GNN_p$) and utilizing only the local GNN channel hierarchy.
% \item \method{}$\neg d$: a variant of \method{} with gradient reverse layer (domain classifier) of \method{} removed.
% \item \method{}$\neg t$: a variant of \method{} which removes the target domain classifier loss of \method{}.
% \end{itemize}

%\begin{figure}[t]
%\centering
%\includegraphics[width=9cm,keepaspectratio=true]{IMG_NEW/Fig.4.pdf}
% \caption{Left: Impact of spectral augmentation radio $\alpha$; Right: Impact of balance ratio $\gamma$ in loss.}
%\caption{Sensitivity analysis on different tasks.}
% \vspace{-0.4cm}
%\label{sensitive}
%\end{figure}

\begin{figure}[t]
\setlength{\abovecaptionskip}{-0.5pt} 
\centering
	\begin{minipage}{0.49\linewidth}
		\centering
            \subfigure{
		\includegraphics[width=1\linewidth]{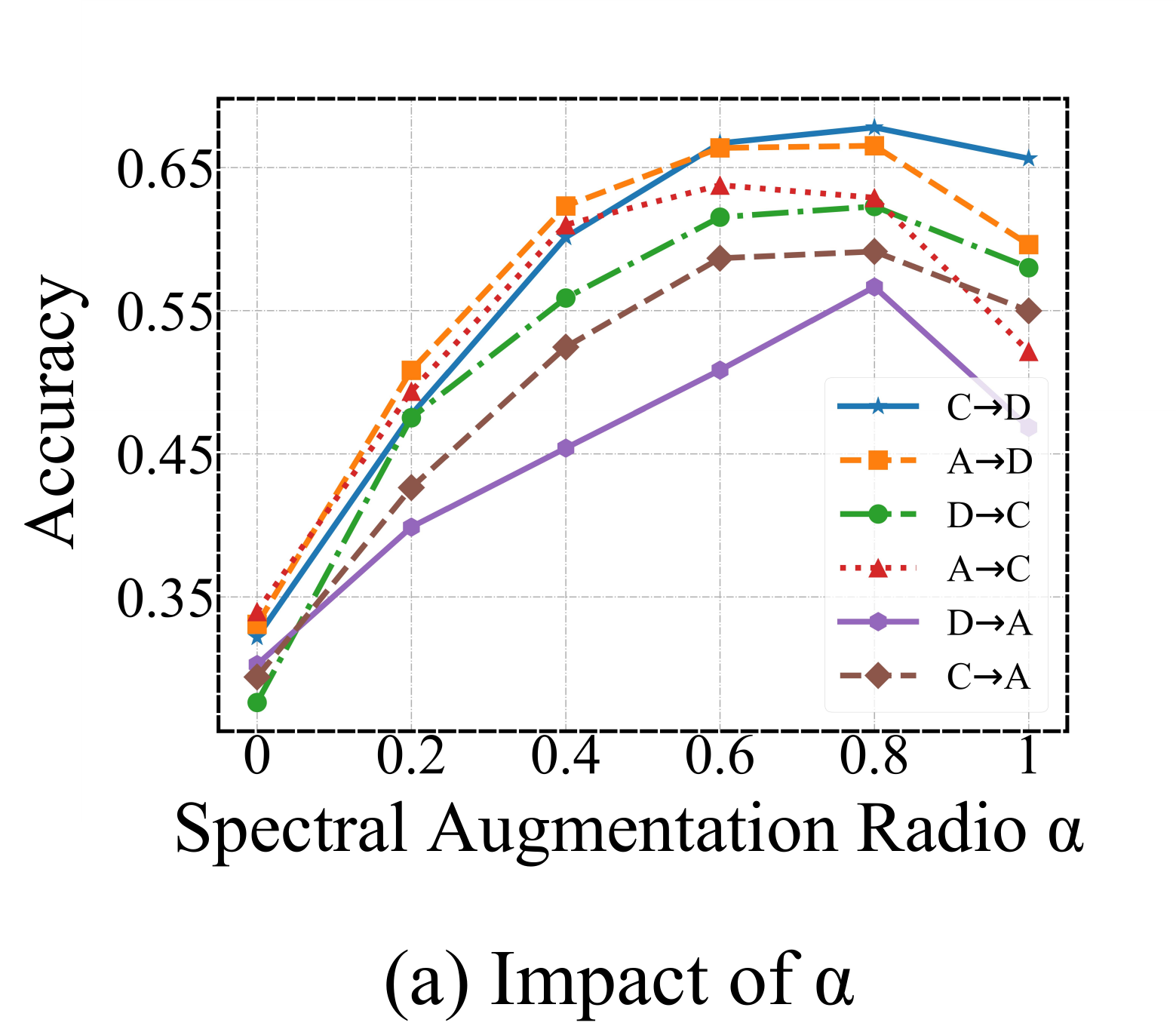}}
	\end{minipage}
	%\qquad
	\begin{minipage}{0.49\linewidth}
		\centering
            \subfigure{
		\includegraphics[width=1\linewidth]{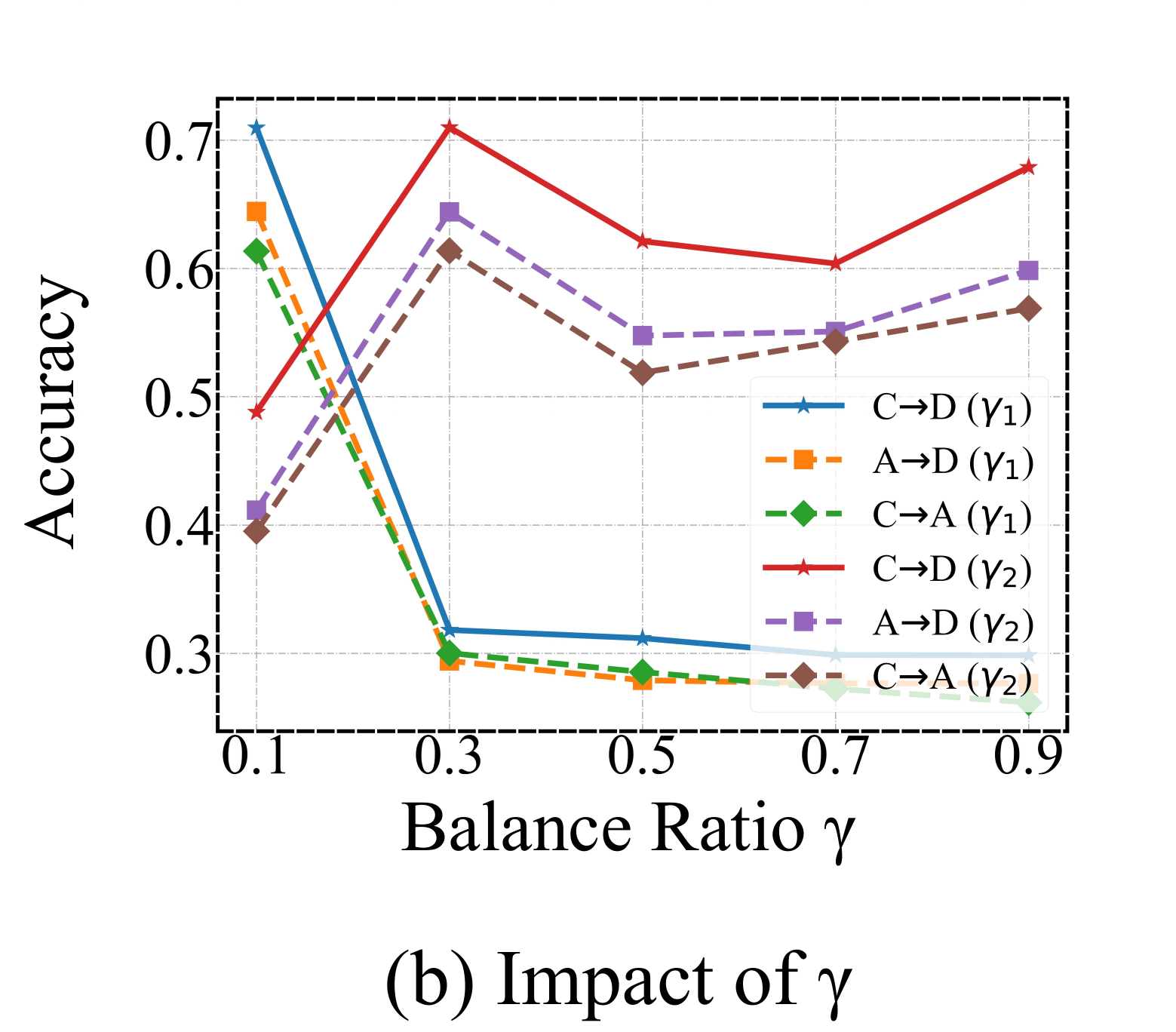}}
	\end{minipage}

        \caption{Sensitivity analysis on different tasks.}

    \label{sensitive}
% \vspace{-0.3cm}
\end{figure}

\begin{figure*}[t]
\centering
\begin{minipage}{4cm}
\centering
\subfigure[DNN]{
\includegraphics[width=1\textwidth]{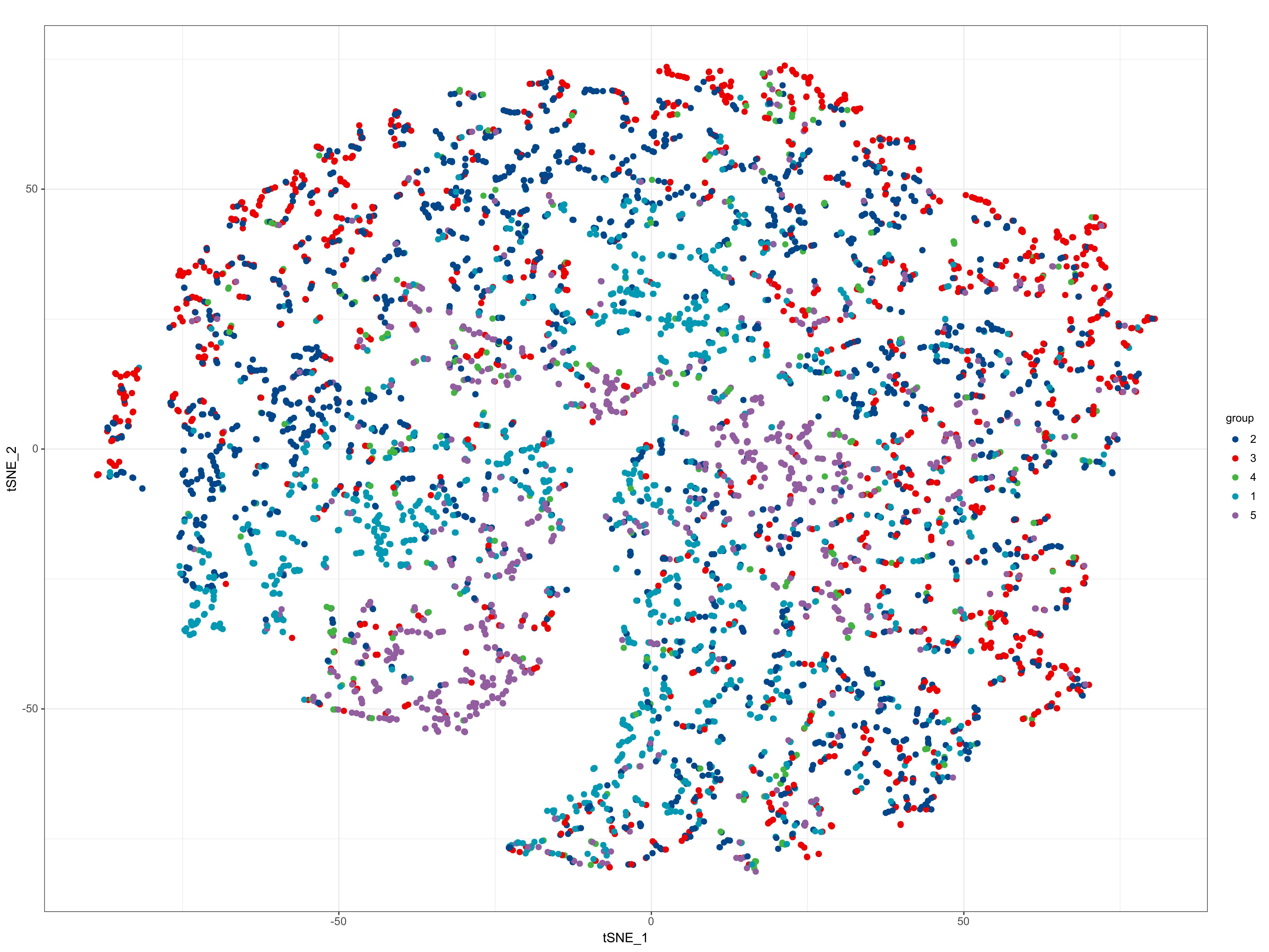}}
\centering
\end{minipage}
\begin{minipage}{4cm}
\centering
\subfigure[GraphSAGE]{
\includegraphics[width=1\textwidth]{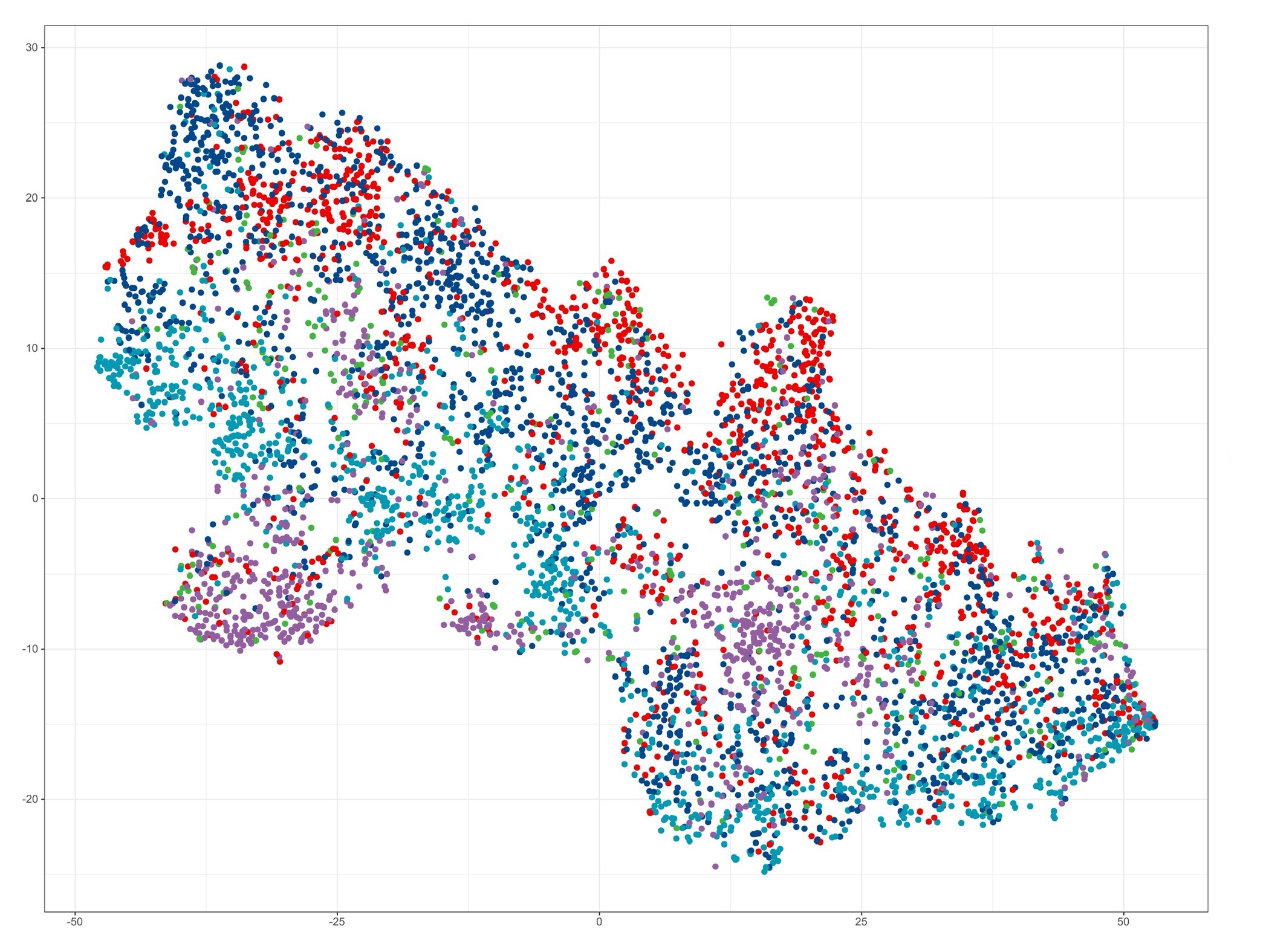}}
\centering
\end{minipage}
\begin{minipage}{4cm}
\centering
\subfigure[UDA-GCN]{
\includegraphics[width=1\textwidth]{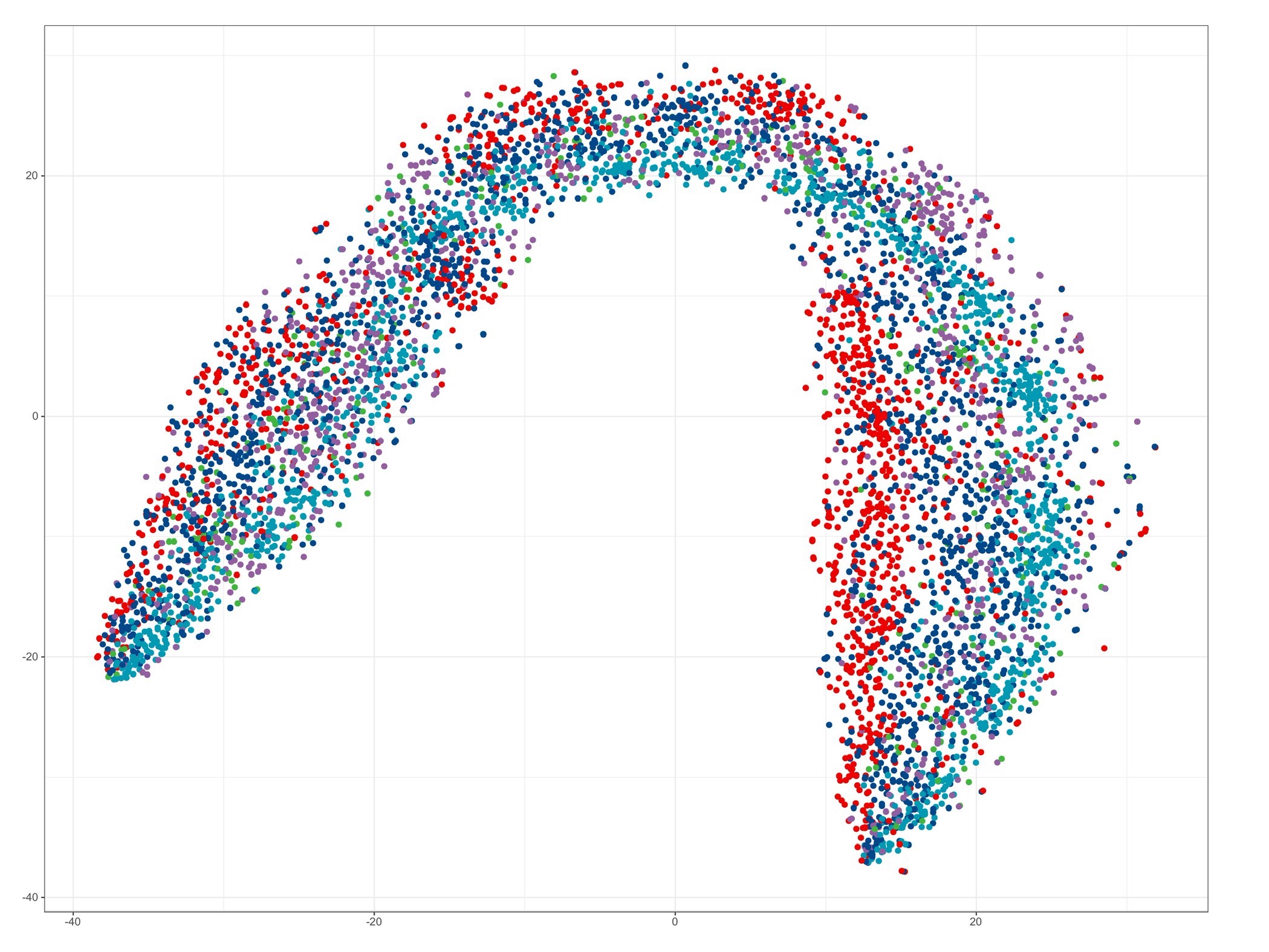}}
\centering
\end{minipage}
\begin{minipage}{4cm}
\centering
\subfigure[\method{}]{
\includegraphics[width=1\textwidth]{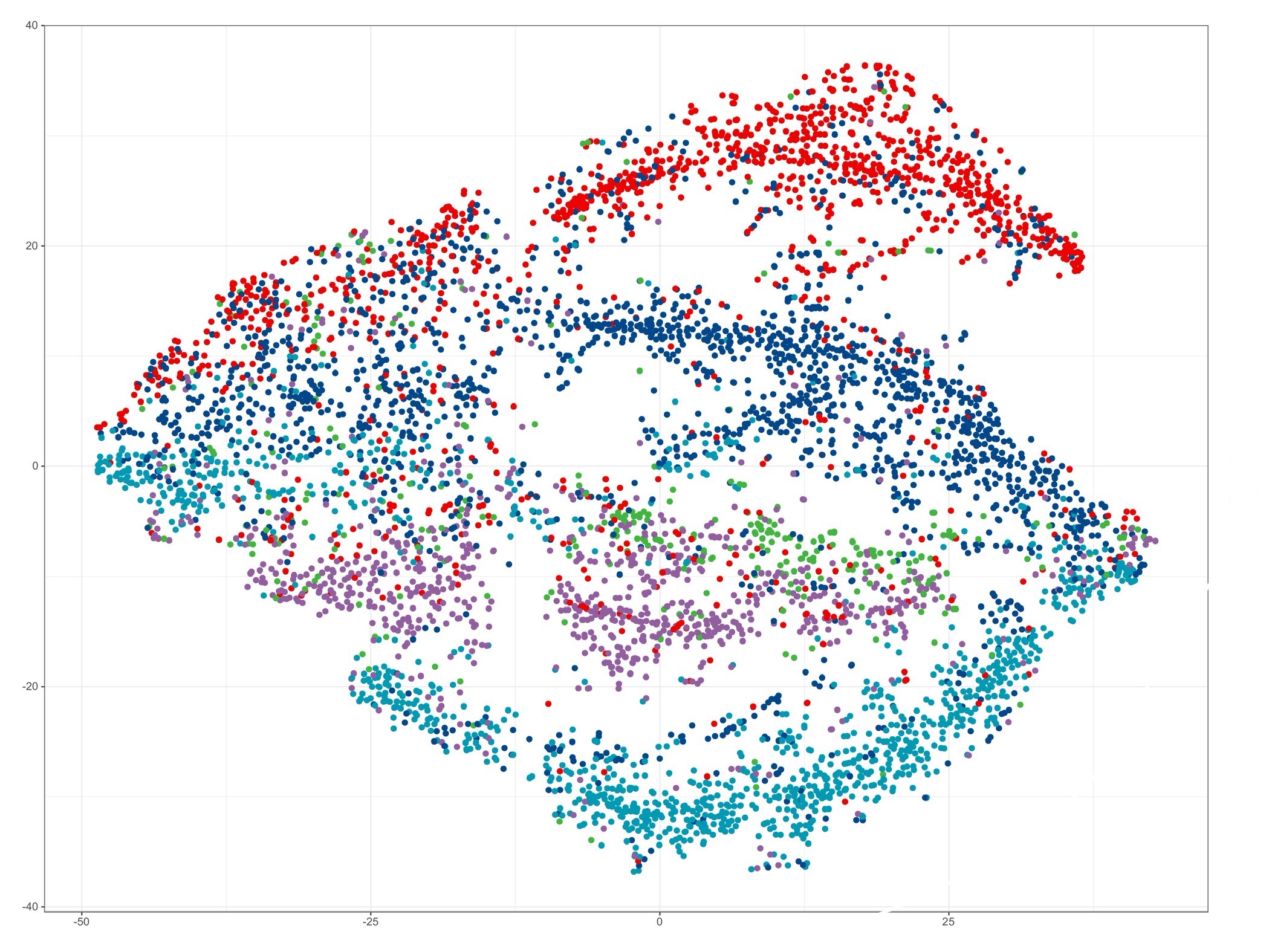}}
\centering
\end{minipage}
\caption{Visualization of the embedding features learned from different methods using $t$-SNE.}
\label{visual}
% \vspace{-0.1cm}
\end{figure*}
% The ablation study results are shown in Table 3.
\begin{itemize}[leftmargin=*]
\item \textit{Impact of Low-frequency Signals}:
To justify the importance of low-frequency signals in spectral domain, we compare the variant model \method{}$\neg l$ with the original model \method{}. As shown in Table~\ref{ablation}, we can evidently observe that the performance of the original model in all tasks are substantially enhanced compared with \method{}$\neg l$. This indicates that the combination of low-frequency signals from both domains greatly reduces the category discrepancy between domains and contributes significantly to the cross-domain graph node classification accuracy.

\item \textit{Impact of High-frequency Signals}:
To present the impact of high-frequency signals in spectral augmentation, we compare the proposed model \method{} with \method{}$\neg h$. Through Table~\ref{ablation}, we find that the performance of \method{} in all tasks are suitably improved compared to \method{}$\neg h$. Although the contribution of combining high-frequency signals in both domains is limited compared to the low-frequency signals, but it is still effective for the cross-domain node classification tasks.

 % The source domain is Citationv1 and the target domain is DBLPv7
\item \textit{Impact of Global GNN}:
To verify the effectiveness of the global consistency information extractor in dual GNN module, we compare \method{} with \method{}$\neg p$. Through the results, we observe that the performance of  \method{} is better than \method{}$\neg p$, indicating that the global information would help to improve the representation ability in dual GNN, thus achieving better results.
% the capability of obtaining a more characterized graph node representation by capturing both global and local consistency information, thus demonstrating the effectiveness of the dual GCN channels.

% \begin{figure*}[t]
% \centering
% \begin{minipage}{4cm}
% \centering
% \subfigure[DNN]{
% \includegraphics[width=1\textwidth]{IMG/DNN10-CA.jpeg}}
% \centering
% \end{minipage}
% \begin{minipage}{4cm}
% \centering
% \subfigure[GraphSAGE]{
% \includegraphics[width=1\textwidth]{IMG/GraphSAGE-CA.jpeg}}
% \centering
% \end{minipage}
% \begin{minipage}{4cm}
% \centering
% \subfigure[UDA-GCN]{
% \includegraphics[width=1\textwidth]{IMG/UDAGCN-CA.jpeg}}
% \centering
% \end{minipage}
% \begin{minipage}{4cm}
% \centering
% \subfigure[\method{}]{
% \includegraphics[width=1\textwidth]{IMG/UDFAGNN-CA.jpeg}}
% \centering
% \end{minipage}
% \caption{Visualization of the embedding features learned from different methods using $t$-SNE\cite{2008Visualizing}. The source domain is Citationv1 and the target domain is ACMv9}
% \end{figure*}

\item \textit{Impact of Domain Adversarial Training}:
To evaluate the effectiveness of domain adversarial training module, we compare \method{} with \method{}$\neg d$. In Table~\ref{ablation}, the performance of \method{} is improved by a certain magnitude compared with \method{}$\neg d$, we contribute the reason to that the presence of domain classifier would reduce the discrepancy between source and target domains, thus increasing the accuracy of the target domain.

\item \textit{Impact of the Target Loss}:
From Table~\ref{ablation}, we can find that the proposed model \method{} performs to a certain extent higher than \method{}$\neg t$ in all six cross-domain node classification tasks, which indicates that the target domain classifier loss is a critical information in the cross-domain problem.
\end{itemize}

\subsection{Sensitivity Analysis (RQ3)}
% 融合比例系数r
To answer \textbf{RQ3}, we conduct experiments to evaluate how the hyper-parameters $\alpha$, $\beta$ and $\gamma_1$, $\gamma_2$ affect the performance of proposed \method{}. $\alpha$ and $\beta$ control how much of the high and low-frequency combined in the spectral augmentation module, and $\gamma_1$, $\gamma_2$ control the balance of target classification and domain adversarial loss. In the implementation, we set $\alpha=\beta$ and vary $\alpha$ in $\{0,0.2,0.4,0.6,0.8,1\}$ with other parameters fixed. Besides, we set $\gamma_1$ and $\gamma_2$ in $\{0.1,0.3,0.5,0.7,0.9\}$ respectively. The results is shown in Fig.~\ref{sensitive}, from the results, we have the following observation:
\begin{itemize}[leftmargin=*]
\item \textit{Impact of Spectral Augmentation Ratio $\alpha$}:
Fig.~\ref{sensitive}(a) shows the impact of $\alpha$, we observe that the classification accuracy improves gradually when $\alpha$ changes from 0 to 0.8, while decrease when $\alpha$ ranges in $\{0.8,1\}$. This is because \method{} needs more spectral information from the source domain for training, while if $\alpha$ is too large, the effectiveness of the spectral augmentation would be weakened, resulting the poor performance. Hence, we set $\alpha$ to 0.8 in our implementation.

% indicating that the more source domain information is incorporated in feature fusion within that range of $r$, the better the classification effect is. However, when $r$ changes from 0.8 to 1, we detect a decrease in the classification accuracy, which indicates that in order to achieve the best results, we need a minor proportion of the target domain signal incorporated into the source domain features, although their mixture proportions are distinct for different source and target domains.

\item \textit{Impact of Balance Ratio $\gamma_1$ and $\gamma_2$}:
We conduct a large number of experiments by setting $\gamma_1$ and $\gamma_2$ in $\{0.1,0.3,0.5,0.7,0.9\}$ respectively, and report the best results as shown in Fig.~\ref{sensitive}(b). From Fig.~\ref{sensitive}(b), we find that, when $\gamma_2$ fixed at 0.1, the performance tends to increase first and then decrease in most cases. The reason is that a small $\gamma_1$ would incorporate the target information for classification while a large $\gamma_1$ may introduce more uncertain information for the scarcity of target labels. Thus, we set $\gamma_1$ to 0.3 as default. Similar to $\gamma_1$, we change the value of $\gamma_2$ with $\gamma_1$ fixed to 0.3. We observe that increasing $\gamma_2$ results in better performance when it is small, indicating that the domain classifier would help to improve the accuracy. However, too large $\gamma_2$ may hurt the performance, which demonstrates that too large of domain loss could harm the discrimination information. Thus, we set $\gamma_2$ to 0.1 as default. 
% 特征维度
% \item \textit{Impact of Feature Dimensions}:
% We set the feature dimension of the output node embedding in the source domain to $Z^s$, and similarly, set that of the output node embedding in the target domain to $Z^t$. Our proposed original model \method{} contains a two-channel GNN module with the structure of 128-16, therefore the feature dimension of the final output node embedding is 16. For the purpose of observing the effect of the output node feature dimension on the cross-domain node classification tasks, we set $d$ to 4, 8, 16, 32, 64 and 128 and perform six such tasks respectively.
% We can observe through Figure 4(b) that when $d$ is changed from 4 to 16, the target domain node classification accuracy has a relatively obvious gradual increase, and when $d$ is changed from 16 to 128, the change of this accuracy gradually tends to be smooth, which indicates that the range of 16 to 128 is a more suitable range of output node feature dimension, and our designed model \method{} is satisfied with this condition.
\end{itemize}

% 可视化
\subsection{Visualization (RQ4)}
% \emph{3) Visualization:}
% A highly important application of network representation is to render meaningful visualizations, i.e., laying out a network in a low-dimensional space.
In order to achieve an intuitive effect of proposed \method{},
we visualize the node representation learned in the target domain to answer \textbf{RQ4}. We visualize the Citationv1$\to$DBLPv7 task and compare it with three models (DNN, GraphSAGE and UDA-GCN), and we have the similar observation on other datasets.
% thus verifying the validity of our proposed model \method{}.
In particular, for each method, we represent the node embeddings with the Stochastic Neighbor Embedding ($t$-SNE)  method, and the results is shown in Fig.~\ref{visual}. From Fig.~\ref{visual},
% The $t$-SNE method maps high-dimensional node embeddings to two- or three-dimensional space for representation, where in large probability, similar embeddings are modeled as closely spaced points, and embeddings with large discrepancies are modeled as distant points.
% Therefore, through the $t$-SNE method, we can derive the similarity relationship between node embeddings in the visualization results.
% Figures 2 and 3 show the visualization results of two groups of target domain node embeddings under different methods, respectively.
we observe that the visualization results of DNN do not have a clear expressive meaning due to the absence of a large number of similar nodes clustered together and the obvious overlap between classes.
Compared with DNN, GraphSAGE and UDA-GCN achieve better visualization representation and the relationship between classes becomes clearer. However, the problem of unclear clustering boundaries still exists.
In the visual representation of our proposed \method{}, the clustering results are clearer compared to the above methods. Meanwhile, there are more obvious boundaries between classes, which indicates that \method{} can generate more meaningful node representation.

%% file: 6.conclusion.tex
\section{Conclusion}

In this paper, we study a less-explored problem of unsupervised domain adaptation for graph node classification and propose a novel framework \method{} to solve the problem. First, we find the inherent features of same category in different domains, i.e., the correlation of  nodes with same category is high in spectral domain, while different categories are distinct. Following the observation, we apply spectral augmentation for category alignment instead of whole feature space alignment. Second, the dual graph extract the local and global information of graphs to learn a better representation for nodes. Last, by using the domain adversarial learning cooperating with source and target classification loss, we are able to reduce the domain discrepancy and achieve the domain adaptation. We conduct extensive experiments on real-world datasets, the results show that our proposed \method{} outperforms the existing cross domain node classification methods. Although the proposed \method{} has achieved impressive results, the complexity of matrix decomposition is high. In the future, we will explore more efficient spectral domain alignment methods and utilize spatial domain features to assist the alignment of category features in different domains, thereby reducing the model complexity. 
% Besides, we will extend \method{} to a broader range of tasks, such as multi-source domain adaptation, and graph noise learning.

%% file: 8.Acknowledge.tex
\section{acknowledge}
This work was supported by the National Key R\&D Program of China under Grant No. 2020AAA0108600.

%% file: 7.Appendix.tex
\clearpage

% \section{Algorithm}
% \label{algorithm}

% \begin{algorithm}[t]
% \caption{Training Algorithm of \method{}}
% \label{alg1}
% \begin{algorithmic}[1]
% \REQUIRE Source domain adjacency matrix $A_s$, degree matrix $D_s$; target domain adjacency matrix $A_t$, degree matrix $D_t$, updating steps $E$; \
% \ENSURE Parameters $\theta$ for the neural network;
% \STATE Initialize $\theta$;
% % \STATE $\mathcal{Q} \leftarrow \emptyset$ ;
% % \REPEAT
% % \REPEAT
% \STATE Transform the source and target spatial graph features into spectral domain.
% \FOR{$e=1,2, \cdots, E$}
% \STATE Calculate the spectral augmentation features on the source domain $Z_s$ with Eq.~\ref{source};
% \STATE Calculate the local and global features with dual graph on the target domain $Z_t$ with Eq.~\ref{target};
% \STATE Update the network parameters $\theta$ through backpropagation by Eq. \ref{total_loss};
% \ENDFOR

% % \UNTIL convergence
% % \STATE Generate each $l_j$;
% % \STATE Cluster the shallow features to get each $e_j$;
% % \STATE Updating $\mathcal{Q}$ by Eq. \ref{eq:Q};
% % \UNTIL convergence
% \end{algorithmic}
% \end{algorithm}

\section{Theoretical Analysis}
In this subsection, we proof the stability of \method{}, which is inspired by~\cite{you2023graph}:
\begin{lemma}
Suppose the $G_s$ and $G_t$ are the graphs of the source and target domain. Given the Laplace matrices decomposition $\bm{L}_d=\bm{D}_d-\bm{A}_d=\bm{U}_d\bm{\Lambda}_d\bm{U}_d^\top$, where $\bm{\Lambda}_d=diag[\lambda_{d1},\cdots,\lambda_{dn}]$ are the sorted eigenvalues of $L_d$, and $d\in\{s,d\}$ denotes the source and target domains. The GNN is constructed as $f(G)=\sigma((g_\theta(\bm{L})\bm{XW})=\sigma(\bm{U}g_\theta(\bm{\Lambda})\bm{U}^\top \bm{XW})$, where $g_\theta$ is the polynomial function with $g_\theta(\bm{L})=\sum_{k=0}^{K}\theta_k \bm{L}^k$, $\bm{W}$ is the learnable matrix and the pointwise nonlinearity has $|\sigma(b)-\sigma(a)|\le |b-a|$. Assuming $||\bm{X}||_{op} \le 1$ and $||\bm{W}||_{op} \le 1$, we have the following inequality:
\begin{equation}
\begin{aligned}
||f(G_s+G_t)-f(G_t)||_2
\le &\alpha [C_\lambda (1+\tau)||\bm{L}_s -\bm{P}^\star \bm{L}_t \bm{P}^{\star \top}||_F\\
&+\mathcal{O}(||\bm{L}_s-\bm{P}^\star \bm{L}_t \bm{P}^{\star \top}||_F^2) \\
&+ max(|g_\theta(\bm{L}_t)|)||\bm{X}_s-\bm{P}^\star \bm{X}_t||_F ],
\end{aligned}
\end{equation}
where $\tau=(||\bm{U}_s-\bm{U}_t||_F+1)^2-1$ stands for the eigenvector misalignment which can be bounded. $\Pi$ is the set of permutation matrices, and $\bm{P}^\star=argmin_{\bm{P}\in\Pi}{||\bm{X}_s-\bm{PX}_t||_F+||\bm{A}_s- \bm{PA}_t\bm{P}^\top||_F}$. $\mathcal{O}(||\bm{L}_s-\bm{P}^\star \bm{L}_t \bm{P}^{\star \top}||_F^2$ is the remainder term with bounded multipliers defined in~\cite{gama2020stability}, and $C_\lambda$ is the spectral Lipschitz constant that $\forall \lambda_i, \lambda_j, |g_\theta (\lambda_i)-g\theta (\lambda_j)|\le C_\lambda (\lambda_i -\lambda_j)$.

\label{prop}
\end{lemma}

\textit{Proof.} Similar to~\cite{you2023graph}, we use $\bm{P}^\star$ as the optimal permutation matrix for $G_s$ and $G_t$. With Eq.~\ref{source}, we have the difference of GNN:
% by combining the high and low-frequency information from source and target domain, which is denoted as $f(G_s+G_t)$, and we have the difference of the GNN outputs:
\begin{equation}
\begin{aligned}
\label{part1}
&||f(G_s+G_t)-f(G_t)||_2 \\
= &||\sigma(\bm{U}_s[\alpha g_\theta^H(\bm{\Lambda}_s)\bm{U}_s^\top \bm{X}_s\bm{W}+(1-\alpha) g_\theta^H(\bm{\Lambda}_t)\bm{U}_t^\top \bm{X}_t\bm{W} \\
&+\beta g_\theta^L(\bm{\Lambda}_s)\bm{U}_s^\top \bm{X}_s\bm{W}+(1-\beta) g_\theta^L(\bm{\Lambda}_t)\bm{U}_t^\top \bm{X}_t\bm{W}]) \\
&- \sigma(\bm{U}_t g_\theta (\bm{\Lambda}_t) \bm{U}_t^\top \bm{X}_t \bm{W})||_2,
% =&||\sigma(U_s[\alpha g_\theta^H(\Lambda_s)U_s^\top X_sW+(1-\alpha) g_\theta^H(\Lambda_t)U_t^\top X_tW \\
% &+\alpha g_\theta^L(\Lambda_s)U_s^\top X_sW+(1-\alpha) g_\theta^L(\Lambda_t)U_t^\top X_tW]) \\
% &- U_t g_\theta (\Lambda_t) U_t^\top X_t W ||_2,
\end{aligned}
\end{equation}
with the triangle inequality and the assumption $|\sigma(b)-\sigma(a)|\le|b-a|, \forall a,b\in\mathbb{R}$, we have:
\begin{equation}
\begin{aligned}
\label{part2}
Eq.\ref{part1}\le & ||\bm{U}_s[\alpha g_\theta^H(\bm{\Lambda}_s)\bm{U}_s^\top \bm{X}_s\bm{W}+(1-\alpha) g_\theta^H(\bm{\Lambda}_t)\bm{U}_t^\top \bm{X}_t\bm{W}  \\
&+\beta g_\theta^L(\bm{\Lambda}_s)\bm{U}_s^\top \bm{X}_s\bm{W}+(1-\beta) g_\theta^L(\bm{\Lambda}_t)\bm{U}_t^\top \bm{X}_t\bm{W}] \\
&- \sigma(\bm{U}_t g_\theta (\bm{\Lambda}_t) \bm{U}_t^\top \bm{X}_t \bm{W})||_F \quad (setting \,\, \alpha=\beta)\\
= &||\bm{U}_s[\alpha (g_\theta^H(\bm{\Lambda}_s)+g_\theta^L(\bm{\Lambda}_s))\bm{U}_s^\top \bm{X}_s \bm{W} \\
&+ (1-\alpha)(g_\theta^H(\bm{\Lambda}_t)+g_\theta^L(\bm{\Lambda}_t))\bm{U}_t^\top \bm{X}_t \bm{W}]\\
&-\bm{U}_t g_\theta (\bm{\Lambda}_t) \bm{U}_t^\top \bm{X}_t \bm{W} ||_F,
\end{aligned}
\end{equation}
where $g_\theta^H$ and $g_\theta^L$ denote the high-pass filter and low-pass filter, which can be designed manually. Setting $g_\theta^H+g_\theta^L=g_\theta$, then Eq.\ref{part2} holds:
\begin{equation}
\begin{aligned}
\label{part3}
Eq.\ref{part2}\le &||\bm{U}_s[\alpha g_\theta(\bm{\Lambda}_s) \bm{U}_s^\top \bm{X}_s \bm{W} + (1-\alpha)g_\theta(\bm{\Lambda}_t) \bm{U}_t^\top \bm{X}_t \bm{W}]\\
&-\bm{U}_t g_\theta (\bm{\Lambda}_t) \bm{U}_t^\top \bm{X}_t \bm{W} ||_F \\
= & ||\alpha \bm{U}_s g_\theta (\bm{\Lambda}_s) \bm{U}_s^\top \bm{X}_s \bm{W} + (1-\alpha) \bm{U}_s \bm{U}_t^{-1} \bm{U}_t g_\theta (\bm{\Lambda}_t) \bm{U}_t^\top \bm{X}_t \bm{W} \\
&-\bm{U}_t g_\theta (\bm{\Lambda}_t) \bm{U}_t^\top \bm{X}_t \bm{W}  ||_F.
\end{aligned}
\end{equation}
$\bm{U}_s$ and $\bm{U}_t$ are the eigen-matrix of $\bm{L}_s$ and $\bm{L}_t$, thus
$\bm{U}_s\bm{U}_s^T=\bm{I}$, $\bm{U}_t\bm{U}_t^T=\bm{I}$. Assuming $\bm{U}_s=[\bm{a}_0,\cdots,\bm{a}_i,\cdots,\bm{a}_n]$ and $\bm{U}_t=[\bm{b}_0,\cdots,\bm{b}_i,\cdots,\bm{b}_n]$, then $||\bm{a}_i||_2=1$ and $||\bm{b}_i||_2=1$, and
$\bm{U}_s\bm{U}_t^{-1}=\bm{U}_s\bm{U}_t^\top=[\bm{a}_0\bm{b}_0^\top,\cdots,\\\bm{a}_i\bm{b}_i^\top,\cdots]\in[-\frac{||\bm{a}_i||_2^2+||\bm{b}_i||_2^2}{2},\frac{||\bm{a}_i||_2^2+||\bm{b}_i||_2^2}{2}]=[-\mathbf{1},\mathbf{1}]$. Therefore:
\begin{equation}
\begin{aligned}
\label{part4}
Eq.\ref{part3} \le & ||\alpha \bm{U}_s g_\theta (\bm{\Lambda}_s) \bm{U}_s^\top \bm{X}_s \bm{W} + (1-\alpha) \bm{U}_t g_\theta (\bm{\Lambda}_t) \bm{U}_t^\top \bm{X}_t \bm{W} \\
&-\bm{U}_t g_\theta (\bm{\Lambda}_t) \bm{U}_t^\top \bm{X}_t \bm{W}  ||_F \\
= & ||\alpha \bm{U}_s g_\theta (\bm{\Lambda}_s) \bm{U}_s^\top \bm{X}_s \bm{W} -\alpha \bm{U}_t g_\theta (\bm{\Lambda}_t) \bm{U}_t^\top \bm{X}_t \bm{W} ||_F \\
=& ||\alpha g_\theta (\bm{L}_s) \bm{X}_s \bm{W} -\alpha g_\theta (\bm{P}^\star \bm{L}_t \bm{P}^{\star \top})\bm{P}^\star \bm{X}_t \bm{W} ||_F.
\end{aligned}
\end{equation}
For any two matrices $\bm{A},\bm{B}$, $||\bm{AB}||_F\le min(||\bm{A}||_{op}||\bm{B}||_F,||\bm{A}||_F||\bm{B}||_{op})$, we have:
\begin{equation}
\begin{aligned}
\label{part5}
Eq.\ref{part4}\le & \alpha ||\bm{W}||_{op} (||g_\theta(\bm{L}_s) \bm{X}_s - g_\theta (\bm{P}^\star \bm{L}_t \bm{P}^{\star \top}) \bm{X}_s \\
&+ g_\theta (\bm{P}^\star \bm{L}_t \bm{P}^{\star \top}) \bm{X}_s - g_\theta (\bm{P}^\star \bm{L}_t \bm{P}^{\star \top})\bm{P}^\star \bm{X}_t ||_F) \\
\le &\alpha ||\bm{W}||_{op} ||\bm{X}_s||_{op} ||g_\theta(\bm{L}_s) - g_\theta (\bm{P}^\star \bm{L}_t \bm{P}^{\star \top})||_F \\
&+ \alpha ||\bm{W}||_{op} ||g_\theta (\bm{P}^\star \bm{L}_t \bm{P}^{\star \top})||_{op} ||\bm{X}_s-\bm{P}^\star \bm{X}_t||_F.
\end{aligned}
\end{equation}
Assuming $||\bm{X}||_{op}\le 1$, $||\bm{W}||_{op}\le 1$ which can be guaranteed by normalization. Besides, learn from~\cite{gama2020stability}, we can get:
\begin{equation}
\begin{aligned}
\label{part6}
Eq.\ref{part5}\le & \alpha ||g_\theta(\bm{L}_s) - g_\theta (\bm{P}^\star \bm{L}_t \bm{P}^{\star \top})||_F \\
&+ \alpha \cdot max(|g_\theta (\bm{L}_t)|) ||\bm{X}_s-\bm{P}^\star \bm{X}_t||_F\\
\le & \alpha [C_\lambda (1+\tau)||\bm{L}_s-\bm{P}^\star \bm{L}_t \bm{P}^{\star \top}||_F+\mathcal{O}(||\bm{L}_s-\bm{P}^\star \bm{L}_t \bm{P}^{\star \top}||_F^2) \\
&+ max(|g_\theta(\bm{L}_t)|)||\bm{X}_s-\bm{P}^\star \bm{X}_t||_F ]. \nonumber
\end{aligned}
\end{equation}
% \end{lemma}
Proof completed.